\documentclass[10pt,conference]{IEEEtran}
\IEEEoverridecommandlockouts
\usepackage{cite}
\usepackage{amsmath,amssymb,amsfonts}
\usepackage{algorithmic}
\usepackage{graphicx}
\usepackage{textcomp}
\usepackage{xcolor}
\usepackage{tcolorbox}
\usepackage{makecell} % 加载makecell宏包
\usepackage{multirow}
\usepackage{subfigure}
\usepackage{booktabs} % 导入 booktabs 宏包
\usepackage{float}
\usepackage{stfloats}
\usepackage[multiple]{footmisc}
\usepackage{tabularx}
\usepackage{marvosym}
\usepackage{arydshln}

\def\BibTeX{{\rm B\kern-.05em{\sc i\kern-.025em b}\kern-.08em
    T\kern-.1667em\lower.7ex\hbox{E}\kern-.125emX}}
\begin{document}

% CoDER: Collaboration of DevOps Experts with Retrieval-augmented Generation
% CoOpERa: Collaboration of AIOps Experts with Retrieval-augmented Generation

\title{CoE-Ops: Collaboration of LLM-based Experts for AIOps Question-Answering\\
% \thanks{Identify applicable funding agency here. If none, delete this.}
}

\author{Jinkun Zhao$^{1}$ \thanks{~\textrm{\Letter}~denotes corresponding author.} 
\quad Yuanshuai Wang$^{1}$ 
\quad Xingjian Zhang$^{1}$ 
\quad Ruibo Chen$^{1}$ 
\quad Xingchuang Liao$^{1}$  
\quad Junle Wang$^{1}$
\\
Lei Huang$^{1,2,~\textrm{\Letter}}$, \quad Kui Zhang$^{1,~\textrm{\Letter}}$, \quad Wenjun Wu$^{1,2,~\textrm{\Letter}}$
\\
$^{1}$SKLCCSE, Institute of Artificial Intelligence,  Beihang University\\
$^{2}$Beijing Advanced Innovation Center for Future Blockchain and Privacy Computing,  Beihang University\\
\\
\normalsize\texttt{\{huangleiai, zhangkui, wwj09315\}@buaa.edu.cn}
}

\maketitle

\begin{abstract}

With the rapid evolution of artificial intelligence, AIOps has emerged as a prominent paradigm in DevOps. Lots of work has been proposed to improve the performance of different AIOps phases. However, constrained by domain-specific knowledge, a single model can only handle the operation requirement of a specific task,such as log parser,root cause analysis. Meanwhile, combining multiple models can achieve more efficient results, which have been proved in both previous ensemble learning and the recent LLM training domain. Inspired by these works,to address the similar challenges in AIOPS, this paper first proposes a collaboration-of-expert framework(CoE-Ops) incorporating a general-purpose large language model task classifier. A retrieval-augmented generation mechanism is introduced to improve the framework's capability in handling both Question-Answering tasks with high-level(Code,build,Test,etc.) and low-level(fault analysis,anomaly detection,etc.). Finally, the proposed method is implemented in the AIOps domain, and extensive experiments are conducted on the DevOps-EVAL dataset. Experimental results demonstrate that CoE-Ops achieves a 72\% improvement in routing accuracy for high-level AIOps tasks compared to existing CoE methods, delivers up to 8\% accuracy enhancement over single AIOps models in DevOps problem resolution, and outperforms larger-scale Mixture-of-Experts (MoE) models by up to 14\% in accuracy.

% DevOps has redefined digital delivery systems through developer-operations collaboration, while the convergence of large-language models and collaboration-of-experts architectures opens new opportunities for intelligent DevOps. However, existing expert collaboration models face two critical limitations in software engineering scenarios: 
% 1) Dependency on domain-specific data for fine-tuning routing models leads to poor generalization and high computational costs; 
% 2) Inability to seamlessly integrate emerging domain-specific expert models. 
% To address these challenges, this paper proposes CoDER, a training-free Collaborative of DevOps Expert Routing framework, featuring two key innovations: 
% 1) Replacing pre-trained classifiers with prompt-engineered general LLMs for cross-phase task intent parsing; 
% 2) Integrating retrieval-augmented generation to mitigate semantic abstraction shifts by dynamically incorporating DevOps knowledge graphs and operational logs. 
% Evaluations on the DevOps-EVAL benchmark demonstrate that CoDER achieves accuracy improvement over state-of-the-art MoE models in deployment orchestration and anomaly attribution tasks, with ablation studies confirming RAG's performance contribution. 
% Requiring no fine-tuning, CoDER enables plug-and-play integration of heterogeneous expert models, providing a scalable solution for LLM-augmented DevOps systems.
\end{abstract}

\begin{IEEEkeywords}
Collaboration of Experts, DevOps, AIOps, Ensemble Learning, Retrieval-augmented Generation.
\end{IEEEkeywords}

\section{Introduction}
% devops -> mlops/aiops -> llmops -> moe/coe
% DevOps, as a revolutionary paradigm, has fundamentally redefined the digital pathway for organizational value creation by bridging the historical divide between software development (Dev) and IT operations (Ops). It establishes a technical delivery system centered on collaborative culture, automated pipelines, and continuous feedback mechanisms. Recent breakthroughs in large language models (LLMs) have catalyzed the emergence of domain-specific expert models tailored for the entire DevOps lifecycle—spanning requirement analysis, code generation, and anomaly detection. These models demonstrate multidimensional task-handling capabilities through context-aware semantic reasoning. Notably, Mixture of Experts (MoE) architectures exemplified by DeepSeek are driving researchers to explore the deep integration of ensemble learning paradigms with LLMs. By leveraging dynamic routing mechanisms to coordinate collaborative decision-making among heterogeneous expert models, this approach provides a novel methodological framework for addressing scalability and robustness challenges in complex DevOps scenarios.

\begin{figure}[t]
\centerline{\includegraphics[width=0.5\textwidth]{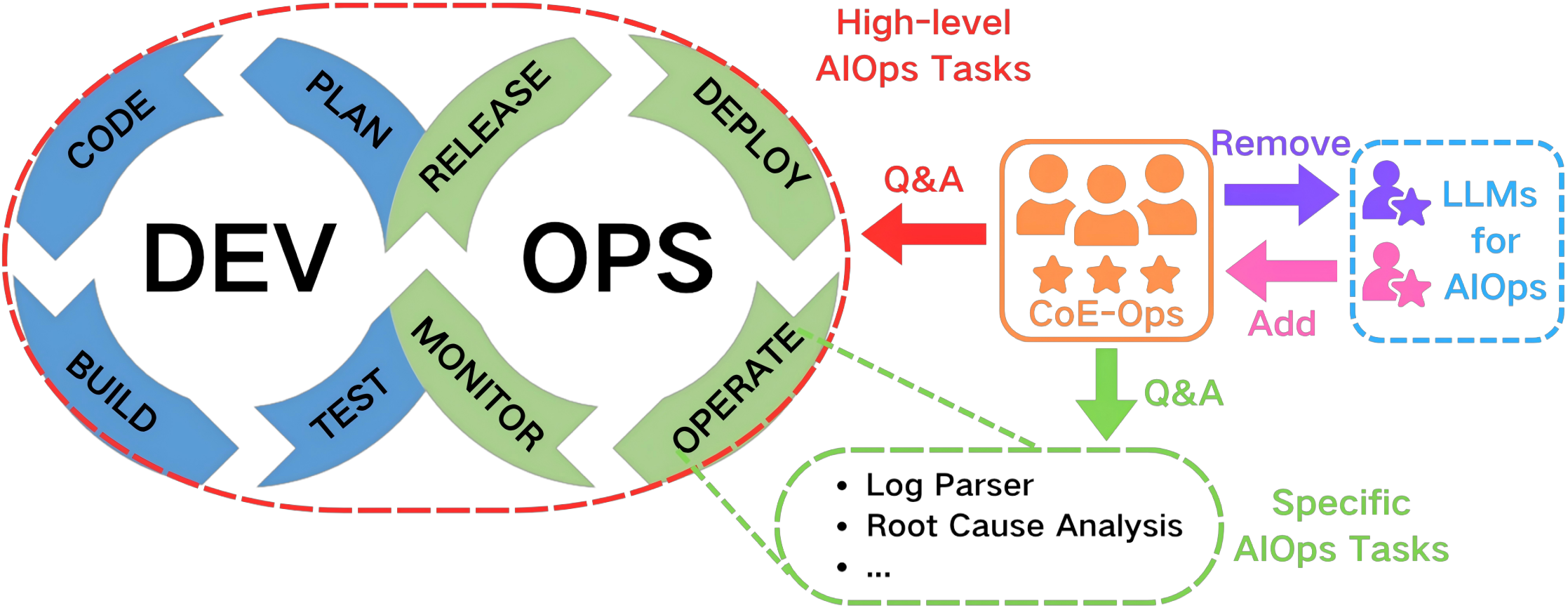}}
\caption{Collaboration Scenarios of CoE-Ops Experts on Question-Answering Tasks at Various Levels within DevOps. CoE-Ops is capable of handling DevOps tasks across the entire life-cycle (high-level) and its sub-tasks (low-level), enabling flexible switching between AIOps experts.}
\vspace{-0.2in}
\label{fig:DEVOPS}
\end{figure}

DevOps is a software engineering methodology designed to bridge the gap between software development (Dev) and IT operations (Ops)\cite{b_d2}. The comprehensive DevOps lifecycle comprises eight iterative phases: Plan, Code, Build, Test, Deploy, Release, Monitor, and Operation. Each phase operates cyclically and encompasses specific subtask categories, as illustrated in Fig.~\ref{fig:DEVOPS}. With advancements in artificial intelligence (AI) and deep learning, emerging paradigms such as MLOps and AIOps have been proposed, representing two distinct approaches to integrating AI with DevOps. Specifically, AIOps employs machine learning models to optimize DevOps workflows\cite{b_d7}\cite{b_d8}\cite{b_d9}.

In recent years, the rapid emergence of large language models has spurred research exploring the integration of LLMs with DevOps. These studies primarily focus on leveraging LLMs to optimize DevOps workflows\cite{b_d13}. Within AIOps implementations that utilize LLMs for DevOps optimization, a critical challenge lies in selecting appropriate AIOps models for different AIOps tasks\cite{b_d1}\cite{b_d9} and enabling expert role-switching capabilities across heterogeneous workflows\cite{b_d3}\cite{b_d7}\cite{b_d16} which is not suitable for multi-agent system leveraging multiple AI agents with specialized and fixed roles\cite{b_a1}. Although domain-specific LLMs tailored for DevOps have emerged, current solutions face limitations due to their reliance on training data from specific domains\cite{b_d8}\cite{b_d11}. This results in inadequate coverage of all DevOps phases and their corresponding subtasks\cite{b_d13}\cite{b_d15}, leading to deployment failures in unfamiliar scenarios\cite{b_d10} and representing a persistent bottleneck for AIOps advancements.

% In recent years, the rapid emergence of large language models (LLMs) has spurred research exploring the integration of LLMs with DevOps. These studies primarily focus on two directions: extending MLOps to include the operational management of LLMs\cite{b_d14}\cite{b_d17}\cite{b_d18}\cite{b_d19}, and further leveraging LLMs to optimize DevOps workflows\cite{b_d13}. Within AIOps implementations that utilize LLMs for DevOps optimization, a critical challenge lies in selecting appropriate models for specific DevOps tasks\cite{b_d1}\cite{b_d9} and enabling role-switching capabilities across heterogeneous workflows\cite{b_d3}\cite{b_d7}\cite{b_d16}. Although domain-specific LLMs tailored for DevOps have emerged, current solutions face limitations due to their reliance on narrow-scoped training data from specific domains\cite{b_d8}\cite{b_d11}. This results in inadequate coverage of all DevOps phases and their corresponding subtasks\cite{b_d13}\cite{b_d15}, leading to deployment failures in unfamiliar scenarios\cite{b_d10} and representing a persistent bottleneck for AIOps advancements.

% To further address the semantic abstraction shift issue in general-purpose large models, we innovatively integrate a Retrieval-Augmented Generation (RAG) mechanism into the CoDER framework. This approach dynamically retrieves domain entities from DevOps knowledge, constructing context-aware prompt vectors to inject fine-grained domain knowledge into the model's reasoning process, thereby mitigating task classification biases. We conduct comprehensive evaluations on DevOps-EVAL, an industry-standard benchmark dataset, employing controlled variable experimental designs to compare against state-of-the-art mixture-of-experts models and collaboration-of-expert frameworks. Ablation studies further quantify the contribution of the RAG component to task classification accuracy (average +18\%). Experimental results demonstrate that CoDER achieves superior scene adaptability and decision interpretability in complex tasks of DevOps, all without requiring fine-tuning.

% To evaluate the performance of the CoE-Ops, This study have formulated the following three research questions (RQs):
Based on the current limitations of AIOps, this paper formulates the following three research questions (RQs):
\begin{itemize}
\item \textbf{RQ1 (Effectiveness):} Can LLM ensembling mitigate the competency gaps between different LLMs?
%  CoE-Ops
\item \textbf{RQ2 (Scalability):} How can LLM ensembling apply for multitask learning in the AIOps domain?

\item \textbf{RQ3 (Efficiency):} Does the integration of smaller models via LLM ensembling enable performance that surpasses that of larger models?

\end{itemize}

To address the current challenges in AIOps regarding model selection\cite{b_d9}, role-switching\cite{b_d16}, and scalability\cite{b_d12}, we propose the following solutions. First, we enhanced the existing Collaboration-of-Experts framework based on a two-stage expert routing mechanism\cite{b_c37}\cite{b_c38}. Subsequently, we integrated the ensemble learning concept and the refined framework into DevOps workflows, leveraging their inherent model-task scalability to enable dynamic selection, composition, and role-switching of AIOps experts (as illustrated in Fig.~\ref{fig:DEVOPS}). Finally, to address high-level DevOps tasks, we incorporated the task classifier with retrieval-augmented generation. This combination enhances task classification by retrieving relevant contextual knowledge for target problems and integrating it into the prompt, thereby improving the framework's adaptability to complex operational scenarios.

% To address the current challenges in AIOps regarding model selection\cite{b_d9}, role-switching\cite{b_d16}, and scalability\cite{b_d12}, we propose the following solutions. First, we enhanced the existing Collaboration-of-Experts framework based on a two-stage expert routing mechanism\cite{b_c37}\cite{b_c38} within ensemble learning methods. This improvement replaces the base models requiring training or fine-tuning with general-purpose large language models integrated with prompt engineering, thereby improving scalability and resulting in our proposed "CoE-Ops" framework. Subsequently, we integrated the ensemble learning concept and the refined framework into DevOps workflows, leveraging their inherent model-task scalability to enable dynamic selection, composition, and role-switching of AIOps experts (as illustrated in Fig.~\ref{fig:DEVOPS}). Finally, to address DevOps tasks characterized by high-level abstraction, we incorporated retrieval-augmented generation. This approach enhances the task classifier’s capabilities by retrieving contextual knowledge relevant to target problems and incorporating it into prompt formulations, thereby further strengthening the framework’s adaptability to complex operational scenarios.

Our key contributions are summarized as follows:
\begin{itemize}
\item A Collaboration-of-Expert framework CoE-Ops based on two-stage expert routing and a general-purpose large language model as task classifier, enabling dynamic switching across diverse AIOps task domains and LLM ensembles.

\item An enhanced task classifier empowered by retrieval-augmented generation technology, specifically designed to address high-level task representations inherent in DevOps scenarios.

\item Comprehensive empirical validation on DevOps-EVAL benchmarks with multiple task-expert configurations and over a dozen AIOps expert models, systematically validating CoE-Ops's dual scalability in task scalability and model scalability.
\end{itemize}

% \begin{itemize}
% \item An enhanced two-stage Collaboration-of-Experts framework CoOpERa is proposed, which replaces conventional trainable task classifiers with general-purpose large language models, thereby eliminating repetitive retraining requirements.

% \item A Retrieval-Augmented Generation mechanism is integ rated into the CoOpERa architecture, substantially improving task resolution capabilities through enhanced knowledge utilization.

% \item Comprehensive domain adaptation and evaluation are conducted by implementing the CoOpERa framework in DevOps scenarios, with systematic validation spanning all operational stages (e.g., development lifecycle, production monitoring) and critical subtasks (e.g., log pattern analysis, infrastructure orchestration), demonstrating cross-domain applicability.
% \end{itemize}

% \begin{itemize}
% \item We propose CoOpERa, an enhanced two-stage Collaboration-of-Experts framework that replaces trainable task classifiers with general-purpose large language models, eliminating the need for repetitive retraining.
% \item We introduce Retrieval-Augmented Generation into the CoOpERa architecture, significantly enhancing its effectiveness in task resolution.
% \item We adapt the CoOpERa framework to the DevOps domain and conduct comprehensive evaluations across all stages (e.g., development, operations) and subtasks (e.g., log parsing, deployment automation), demonstrating its broad applicability.
% \end{itemize}

\section{Related Work}
\subsection{Development and Operations}
\label{subsec:Development and Operations}

DevOps is a collaborative, cross-domain software development methodology that emphasizes the automation of continuous delivery for software updates\cite{b_d3}. When integrated with artificial intelligence, its evolutionary trajectory bifurcates into two primary branches: MLOps and AIOps\cite{b_d12}.

\paragraph{MLOps} MLOps focuses on applying DevOps practices to machine learning systems, aiming to establish seamless integration between diverse open source tools to enable fully automated execution of ML workflows, spanning dataset construction, model training, and deployment\cite{b_d4}\cite{b_d5}. With the recent emergence of large language models, LLMOps\cite{b_d14}, an extension of MLOps tailored for LLM development and deployment, have gained momentum. LLMOps addresses the unique operational challenges of LLMs\cite{b_d17} and provides specialized tools for efficient data processing, model training, deployment, and maintenance\cite{b_d19}. However, both MLOps and LLMOps currently face limitations, including a lack of standardized practices, difficulties in maintaining model consistency and scalability\cite{b_d6}\cite{b_d18}, and ambiguous evaluation criteria.

\paragraph{AIOps} In contrast, AIOps leverages AI and ML technologies to efficiently build and operate large-scale online services and applications in software engineering\cite{b_d7}. Most existing AIOps implementations rely on data from a limited number of domains\cite{b_d8} and predominantly employ supervised learning techniques\cite{b_d11}. Consequently, their proposed models are often confined to specific DevOps subdomains rather than being deployable across the entire ecosystem\cite{b_d13}. A critical challenge for AIOps lies in selecting and integrating appropriate machine learning models\cite{b_d9}\cite{b_d16} to ensure adaptability to diverse use cases while fulfilling heterogeneous\cite{b_d15} and evolving requirements\cite{b_d12}.

\subsection{Ensemble Learning with Large Language Models}
\label{subsec:Ensemble learning}
Ensemble learning with large language models involves the systematic utilization of multiple LLMs, each designed to handle user queries during downstream inference to capitalize on their individual strengths\cite{b_c1}\cite{b_c2}. Depending on the strategy for model integration, ensemble learning can be categorized into two paradigms: Mixture-of-Experts (MoE) and Collaboration-of-Experts (CoE). 

\paragraph{Mixture-of-Experts} In recent years, MoE models have become a primary choice for foundation models\cite{b_c39}\cite{b_c40} due to their computational efficiency and strong generalization capabilities. In MoE systems, different expert modules possess distinct strengths, making efficient utilization a key challenge. FrugalGPT\cite{b_c3} and LLM-Blender\cite{b_c4} aggregate outputs from various experts to generate final results, while others adopt voting strategies to select the optimal output\cite{b_c7}\cite{b_c9}\cite{b_c10}. However, these expert modules cannot complete tasks independently, and the selection and generation processes lack interpretability. As a result, the Collaboration-of-Experts framework has increasingly drawn attention from researchers.

\begin{figure}[t]
\centerline{\includegraphics[width=0.5\textwidth]{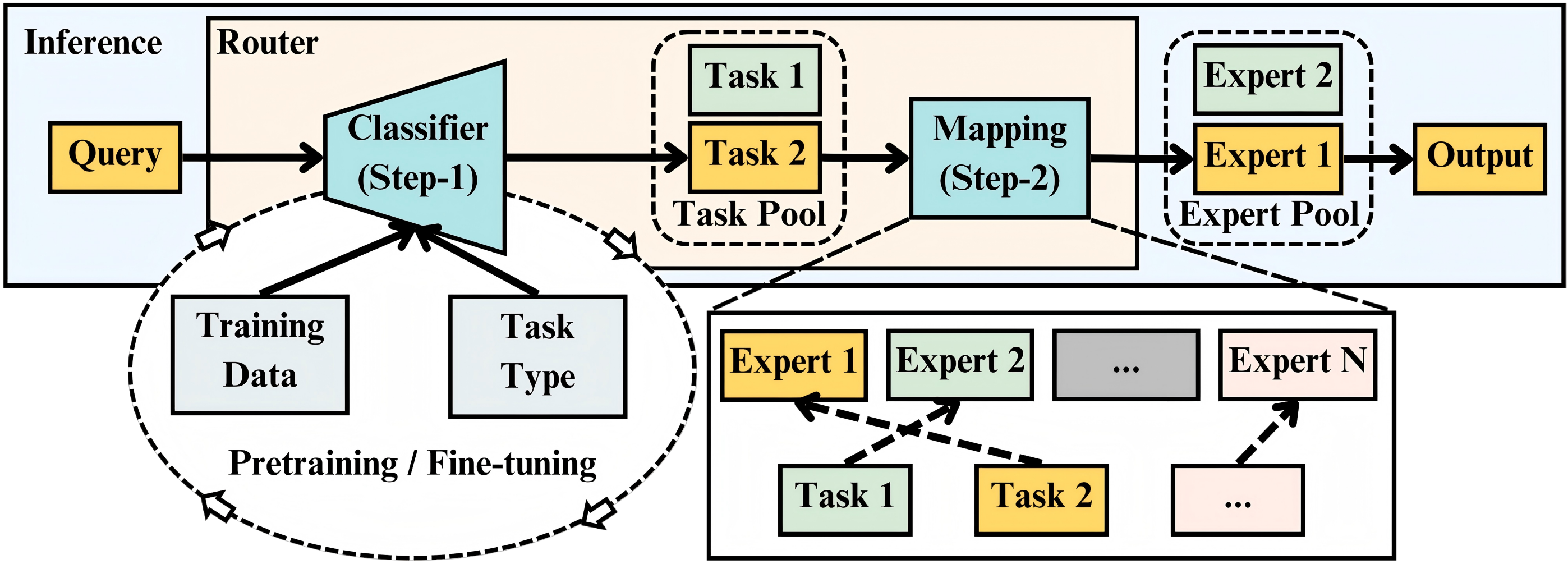}}
\caption{Framework of the CoE Based on Two-Stage Expert Routing. The input is first processed by a pre-trained task classifier to obtain its corresponding task category label (Step-1). It is then routed to a designated expert model based on a pre-established "Task-Expert" mapping derived from existing benchmark (Step-2). Finally, it is the produce by the designated expert model.}
\label{fig:framework_origin}
\vspace{-0.2in}
\end{figure}

\paragraph{Collaboration-of-Experts} CoE primarily facilitates synergistic interactions among experts by selecting one or several optimal experts for a given input. Early efforts explored the use of sub-networks as expert models\cite{b_c11}\cite{b_c15}. With the proliferation of large-scale models, CoE has shifted focus toward incorporating diverse performance metrics, such as answer accuracy\cite{b_c13}\cite{b_c16}, inference cost\cite{b_c12}\cite{b_c17}\cite{b_c18}, and problem difficulty\cite{b_c14}\cite{b_c19}. A core research direction in CoE involves the design of routing algorithms for large models\cite{b_c13}. For instance, cascading networks\cite{b_c24}\cite{b_c28} have been proposed, or large models are represented as nodes\cite{b_c22}\cite{b_c27} or vector embeddings\cite{b_c30}, with probabilistic methods\cite{b_c25} employed to predict routing outcomes. Recent studies further integrate reinforcement learning\cite{b_c31}\cite{b_c32}\cite{b_c33}\cite{b_c34} to refine expert routing strategies and introduce hardware-aware optimizations\cite{b_c35}\cite{b_c36} for efficient expert model loading. To address the lack of interpretability in routing decisions, a two-stage expert routing framework\cite{b_c37}\cite{b_c38} has been developed (as shown in Fig~\ref{fig:framework_origin}). This framework first categorizes input problems and then selects the most suitable expert for each category, thereby enhancing both the explainability of routing decisions and the scalability of the overall system.

% Notably, researchers are increasingly exploring two-stage routing frameworks to enhance the generalization of large language model (LLM) routing systems. For instance, [37] introduced a two-stage routing paradigm:
% A category router first classifies inputs into distinct task categories. A category-to-expert mapper then assigns each category to a specialized expert.This work also proposed a two-phase training protocol—initially training the category router, followed by optimizing the category-to-expert mapping. Similarly, [38] developed a simplified mapping strategy that leverages existing benchmark rankings to infer expert model capabilities and construct category-expert associations.

% \section{Model and Task Scalability}
% \label{subsec:Model and Task Scalability}

\section{Problem Formulation}
\label{sec:Basic Concepts}

\begin{figure}[t]
\centerline{\includegraphics[width=0.5\textwidth]{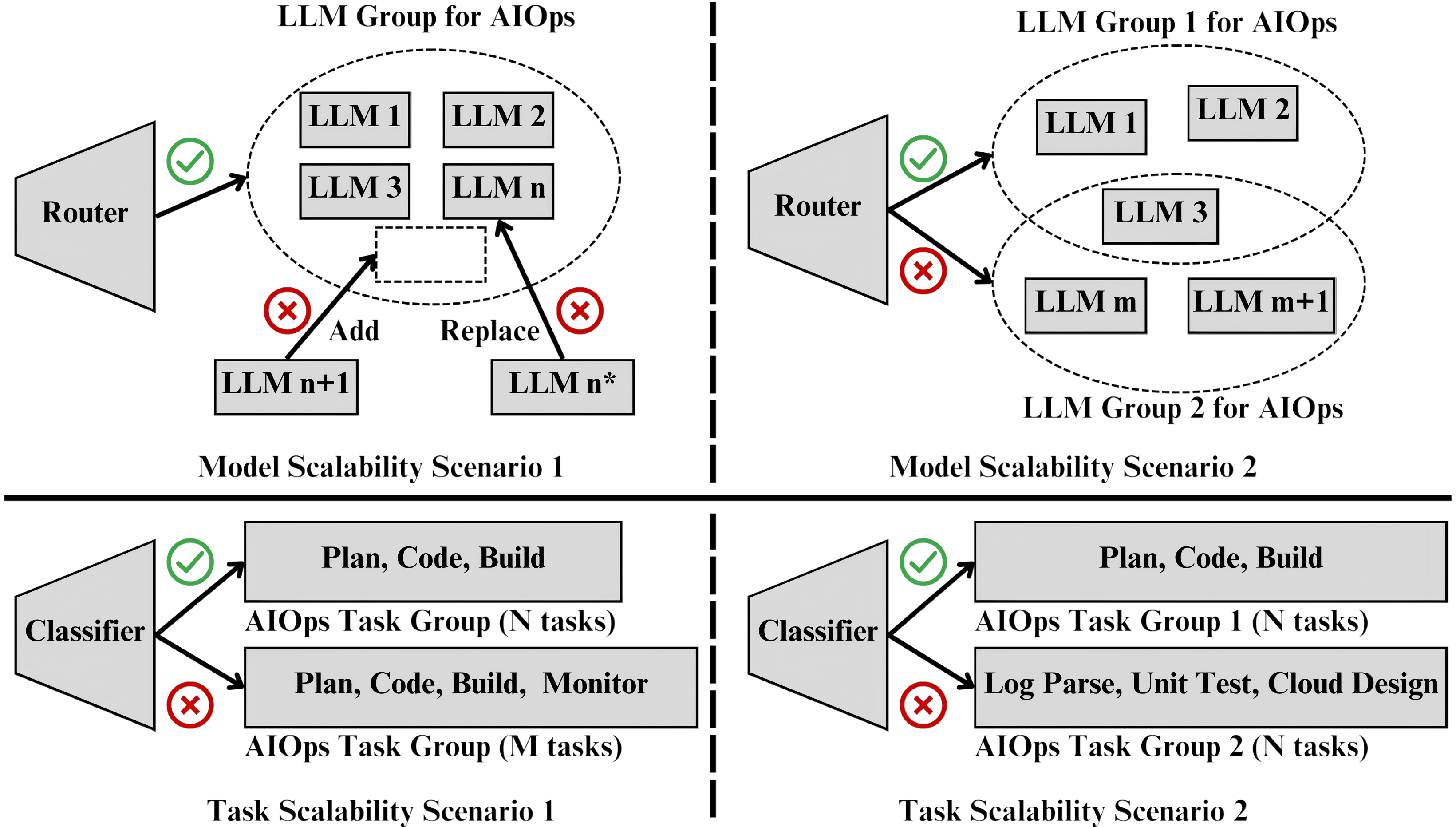}}
\caption{AIOps Scenario Illustrating Model and Task Scalability. The end-to-end expert router is primarily constrained by model scalability. While the classifier in the two-stage expert routing approach improves model scalability, it is still constrained by task scalability.}
\label{fig:model_and_task}
\vspace{-0.2in}
\end{figure}

Before introducing the collaboration-of-experts paradigm into the AIOps domain, it is essential to delineate both the existing challenges within AIOps and the potential limitations this collaborative approach may encounter when addressing highly abstract AIOps tasks. The primary challenge in contemporary AIOps lies in effectively orchestrating diverse LLMs from distinct domains to address multifaceted operational requirements. For collaboration-of-experts frameworks, their scalability emerges as a critical concern within AIOps due to the field’s broad spectrum of tasks and model heterogeneity. We categorize this scalability challenge along two dimensions: model scalability and task scalability. The corresponding operational scenarios for these dimensions are proposed in Fig.~\ref{fig:model_and_task}.

% Most existing methods employ models trained or fully fine-tuned on domain-specific datasets as routers for MoE or CoE systems. While these approaches ensure relatively accurate input-to-expert routing within their target task domains, they suffer from substantial training overheads, dependency on domain-specific datasets, and inherent scalability limitations. To address this, we categorize scalability into two dimensions: model scalability (extending to larger or distributed architectures) and task scalability (adapting to emerging task categories). Corresponding operational scenarios for these dimensions are proposed and visualized in Fig.~\ref{fig:model_and_task}.

\subsection{Model Scalability}
\label{subsec:Model Scalability}
For collaboration-of-experts with end-to-end expert routing, the scalability of model remains a critical challenge. As illustrated in Scenario 1, because the router directly employs LLMs as routing nodes, newly released AIOps LLMs cannot be dynamically incorporated into or replace old models in the router's LLM group. To address this issue, routers must undergo retraining whenever LLM group for AIOps are updated, incurring substantial computational overhead.

Furthermore, Scenario 2 demonstrates that when task contexts evolve, certain models in the existing group may become unsuitable as experts for emerging tasks. However, the router cannot transit to new expert groups tailored to the updated task requirements since it rigidly maps inputs to fixed AIOps experts. This necessitates costly retraining or fine-tuning of the router to adapt to new AIOps task scenarios.

With the emergence of collaboration-of-experts frameworks with two-stage expert routing such as Bench-CoE\cite{b_c38} and Composition of Experts\cite{b_c37}, CoE can now dynamically reconfigure candidate AIOps LLMs through flexible mapping adjustments, thereby enhancing the scalability of model.

\subsection{Task Scalability}
\label{subsec:Task Scalability}
While two-stage expert routers\cite{b_c37}\cite{b_c38} outperform their end-to-end counterparts in model scalability, they still exhibit significant limitations in task scalability. As demonstrated in Scenario 1, the output dimensionality of the two-stage expert routing remains fixed, since it employs discriminative models as classifiers. Consequently, when the number of classification tasks changes (from N to M), structural modifications and retraining are typically required.

Furthermore, Scenario 2 reveals that when task contexts evolve, the task classifier often fails to generalize to unseen tasks without retraining on in-domain data. This limitation stems from the classifier’s reliance on parametric knowledge (memorized during training) rather than leveraging external knowledge sources, restricting its task scalability.

\section{Methodology}

% \begin{figure}[htbp]
% \centerline{\includegraphics[width=0.5\textwidth]{figs_framework/Framework_CoDER.png}}
% \caption{General Framework of CoE-Ops.}
% \label{fig:framework_CoE-Ops}
% \end{figure}

\begin{figure}[t]
\centerline{\includegraphics[width=0.5\textwidth]{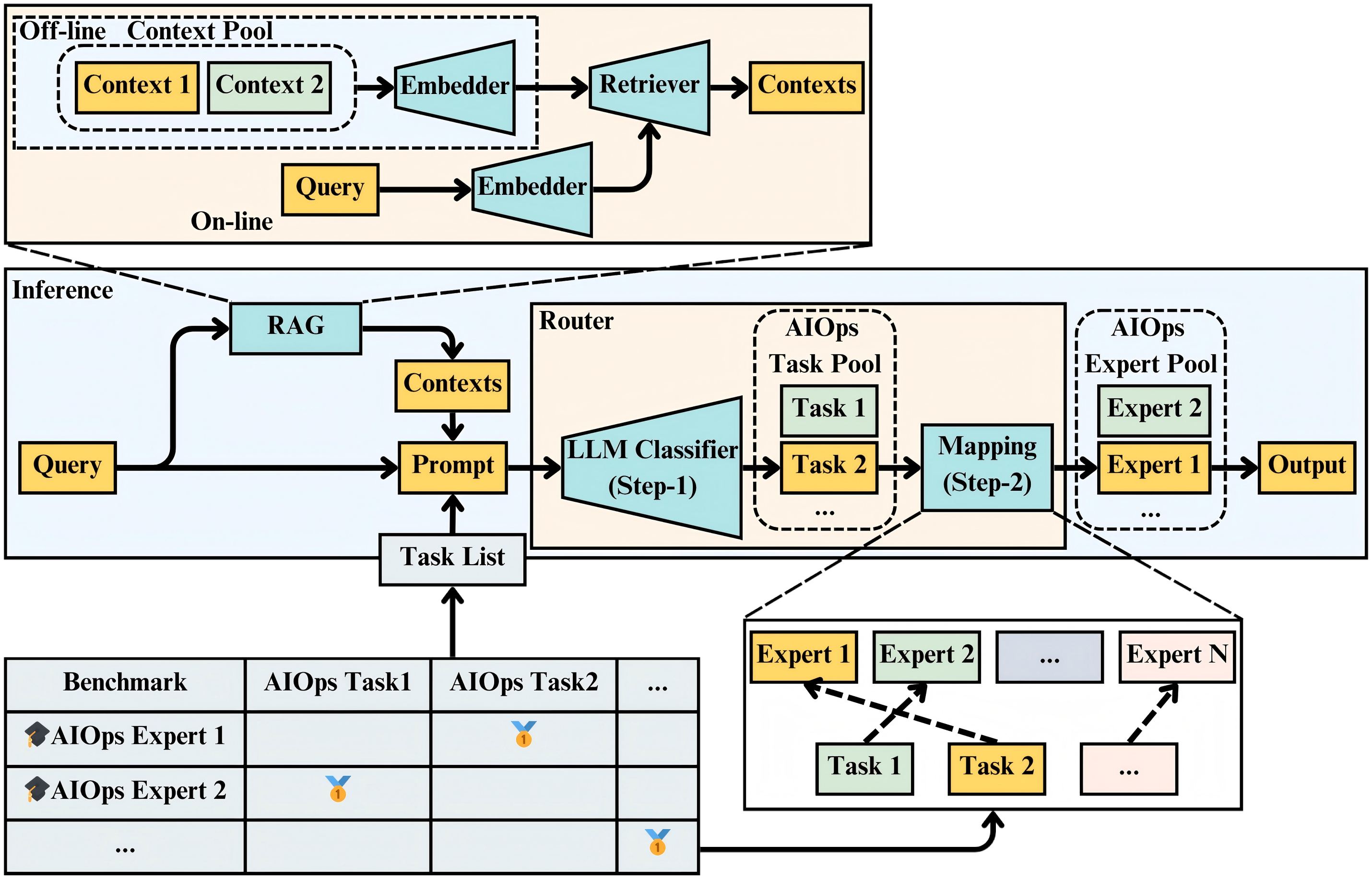}}
\caption{Framework of CoE-Ops. CoE-Ops introduces improvements to Step-1 of the CoE based on two-stage expert routing. First, the discriminative model-based classifier is replaced with an LLM-based classifier. Subsequently, the prompt is enhanced by extracting a task list from benchmark datasets and employing Retrieval-Augmented Generation (RAG) technology to retrieve relevant context for the current input, thereby assisting the LLM-based classifier in classification.}
\label{fig:CoE-Ops with RAG}
\vspace{-0.2in}
\end{figure}

The framework of our proposed CoE-Ops is shown in Fig. \ref{fig:CoE-Ops with RAG}. It consists of a two-stage expert routing mechanism which replaces discriminative models with general-purpose LLMs enhanced by retrieval-augmented generation capabilities.

\subsection{Two-stage Expert Routing}

CoE-Ops primarily improves upon the two-stage expert routing mechanism proposed in seminal works including Composition of Experts\cite{b_c37} and Bench-CoE\cite{b_c38}. During the original process of two-stage expert routing, the AIOps user's query is first classified by a pretrained or fine-tuned classifier to determine its task type. The query is then routed to the best-in-domain model for processing based on this label, as shown in Fig.~\ref{fig:framework_origin}.

The task classifier in the two-stage expert routing can be abstracted as \eqref{eq:Classifier} shows. 
% Be sure that the 
% symbols in your equation have been defined before or immediately following 
% the equation. Use ``\eqref{eq}'', not ``Eq.~\eqref{eq}'' or ``equation \eqref{eq}'', except at 
% the beginning of a sentence: ``Equation \eqref{eq} is . . .''
% classifier
\begin{equation}
\hat{T} = \mathop{\arg\max}\limits_{T \in \{T_1, T_2, \dots, T_n\}} P(T|X, \mathcal{C}),
\label{eq:Classifier}
\end{equation}
where T represents the AIOps task, $\mathcal{C}$ represents the classification model, and X denotes the current input from user.

In particular, within the two-stage expert routing architecture of the Collaboration of Experts, the cardinality of candidate AIOps experts should adhere to the bounds specified in \eqref{eq:expert number}, since each AIOps expert model demonstrates expertise in a minimum of one specialized AIOps domain.
\begin{equation}
2 \leq N_{\text{expert}} \leq N_{\text{task}},
\label{eq:expert number}
\end{equation}
where $N_{\text{task}}$ denotes the number of AIOps tasks.

Following AIOps task categorization by the classifier, input AIOps queries are dynamically routed to domain-specialized expert models through a "task-expert" allocation mechanism, as mathematically formalized in \eqref{eq:mapping}.

% Once the task classifier assigns a task label to the input query, the query is assigned to a designated expert via a "task-expert" mapping, which can be represented by \eqref{eq:mapping}.

% mapping
\begin{equation}
f:\mathcal{T}\to\mathcal{E},
\label{eq:mapping}
\end{equation}
where $\mathcal{T}=\{T_1,T_2,\ldots,T_M\}$ denotes the set of AIOps tasks, $\mathcal{E}=\{E_1,E_2,\ldots,E_N\}$ denotes the set of AIOps experts, $M$ indicates the count of AIOps tasks, and $N$ indicates the count of AIOps experts.

% \begin{equation}
% \text{Benchmark}(M, \mathcal{T}) = \frac{1}{M} \sum_{i=1}^{M} \frac{1}{N_i} \sum_{j=1}^{N_i} \mathbb{I}(M(\mathcal{I}_{ij}) = \mathcal{L}_{ij})
% \end{equation}

When developing the "task-expert" allocation mechanism, it is necessary to establish a metric for evaluating the capability of each expert model across different task domains. For AIOps queries involving multiple-choice questions and question-and-answer formats, the answer accuracy of the expert model can serve as a suitable evaluation metric. This accuracy measurement, as shown in \eqref{eq:accuracy}, provides a quantitative basis for assessing model performance.

\begin{equation}
\text{Accuracy}(M, T_i) = \frac{1}{N_i} \sum_{j=1}^{N_i} \mathbb{I}(M(\mathcal{X}_{ij}) = \mathcal{A}_{ij}),
\label{eq:accuracy}
\end{equation}
where $N_i$ denotes the number of AIOps queries in the AIOps task $T_i$, $M$ represents the expert model, $\mathcal{X}_{i}$ stands for the AIOps queries in the AIOps task $T_i$, and $\mathcal{A}_{ij}$ indicates the correct answer to the AIOps query $\mathcal{X}_{ij}$.

Upon construction of the capability assessment leaderboard, the expert model demonstrating superior accuracy within each task domain is designated as the optimal solution for the "task-expert" allocation, with formal validation provided in \eqref{eq:benchmark}.
%  and \eqref{eq:mapping result}
\begin{equation}
M_i^* = \mathop{\arg\max}\limits_{M \in \mathcal{M}} \left( \frac{1}{N_i} \sum_{j=1}^{N_i} \mathbb{I}(M(\mathcal{X}_{ij}) = \mathcal{A}_{ij}) \right),
\label{eq:benchmark}
\end{equation}
where $M_i^*$ denotes the best AIOps model on task $T_i$.

% \begin{equation}
% f(T_i) = M_i^*
% \label{eq:mapping result}
% \end{equation}

% \begin{tcolorbox}
% this is a box
% \end{tcolorbox}

\subsection{Classifier with General-purpose LLM}
\label{subsec: General Large Language Model}

\begin{tcolorbox}[title = {Prompt 1 - Classifier with General-purpose LLM}]
\label{box: classifier prompt}
You are a classifier that can categorize questions into specific tasks. Your job is to analyze the following given question and determine which task from the provided list it most likely belongs to. \\
The tasks are as follows: \{$task~list$\}. \\
The question is: \\
"\{$question$\} \\
A.\{$option_A$\} \\
B.\{$option_B$\} \\
C.\{$option_C$\} \\
D.\{$option_D$\}".\\
Provide your answer in the format: "**Task: [$selected~task$] **".
\end{tcolorbox}

To overcome the limitations inherent in conventional two-stage expert routing CoE frameworks, particularly their dependence on repeated classifier fine-tuning or retraining across distinct task scenarios, we implement a dual enhancement strategy. First, the classifier component is replaced by a general-purpose LLM operating in zero-shot mode, thereby eliminating fine-tuning requirements. Second, a structured task-list prompting mechanism (see Prompt 1) is integrated to ensure task scalability of the optimized architecture.

% To address the limitations of the current two-stage LLM routing method, which requires repeated fine-tuning of retraining of the classifier for different task scenarios, we incorporated a general-purpose LLM to replace the classifier component without fine-tuning. Furthermore, to enhance the method's scalability across different tasks, we introduced a prompt incorporating a task list, as demonstrated in Prompt 1.

The enhanced framework enables dynamic adaptation to shifting task scenarios through prompt-based task list modification, eliminating the need for classifier pretraining or fine-tuning. This architectural innovation substantially reduces computational overhead while maintaining task scalability within the CoE paradigm.

% In this improved framework, when the task scenario changes, users can switch the task prediction domain of the CoE framework by simply modifying the task list in the prompt, without re-pretraining or fine-tuning the classifier or router. This significantly reduces the training overhead of the framework.

The classification architecture of our framework, enhanced through the integration of prompt engineering and a general-purpose LLM, achieves formal abstraction as mathematically characterized in \eqref{eq:Prompt_Classifier}.
% After incorporating prompts and a general-purpose LLM, our CoDER's classification model can be further abstracted as shown in \eqref{eq:Prompt_Classifier}.

\begin{equation}
\hat{T} = \mathop{\arg\max}\limits_{T \in \{T_1, T_2, \dots, T_n\}} P(T|X, P, \mathcal{L}_{\text{General}}),
\label{eq:Prompt_Classifier}
\end{equation}
where $P$ denotes the prompt with the task list, $\mathcal{L}_{\text{General}}$ represents the general-purpose LLM.

% classify numbers
Notably, unlike fine-tuned classifiers, using a general-purpose LLM as a classifier may yield an "unknown" class result. This reflects the LLM's effort to reduce hallucination by refusing to force-classify ambiguous inputs. Thus, after incorporating prompts and a general-purpose LLM, an additional "unknown" class is needed. Consequently, the number of output task classes is modified as shown in \eqref{eq:Task Number}.
\begin{equation}
N_{\text{predict~task}} = N_{\text{task}} + N_{\text{unk}},
\label{eq:Task Number}
\end{equation}
where $N_{\text{unk}}$ denotes the number of tasks of unknown types (typically equals 1).

In this case, we need to select an extra expert model for the "unknown" class. Our selection strategy, as shown in \eqref{eq:Unknown select}, is to choose the expert model with the highest average capability in all task domains to handle the "unknown" AIOps input.
%  and \eqref{eq:Unknown mapping}
\begin{equation}
M_{\text{unk}}^* = \mathop{\arg\max}\limits_{M \in \mathcal{M}} \left( \frac{1}{N_{\text{total}}} \sum_{T_i \in \mathcal{T}} \sum_{j=1}^{N_i} \mathbb{I}(M(\mathcal{X}_{ij}) = \mathcal{A}_{ij}) \right),
\label{eq:Unknown select}
\end{equation}
where $M_{\text{unk}}^*$ denotes the best AIOps model on "unknown" task.

% \begin{equation}
% f(T_{\text{unknown}}) = M_{unk}^*
% \label{eq:Unknown mapping}
% \end{equation}
\begin{tcolorbox}[title = {Prompt 2 - AIOps Experts with Chain of thought}]
Please answer the following DEVOPS question. \\
The question is: \{$question$\} \\
The options are as follows: \\
A. \{$option_A$\} \\
B. \{$option_B$\} \\
C. \{$option_C$\} \\
D. \{$option_D$\} \\
Think step by step and then finish your answer with "the answer is (X)" where X is the correct letter choice.
\end{tcolorbox}

For the expert models, we also avoid fine-tuning. Instead, we use prompts with chain of thought as the input. The prompt template is shown in Prompt 2. In the multiple-choice setting, to assess expert capabilities via answer accuracy, we ask the model to return answers in a fixed format.

\subsection{LLM Classifier Enhanced with RAG}
\label{subsec: Large Language Model Enhance with RAG}
Simply replacing the classifier in the two-stage expert router with a general-purpose LLM carries risks. In AIOps domains with abstract or high-level task (like plan, build, code, etc.), the LLM may struggle to link inputs to tasks due to limited information. To address this, context needs to be introduced to help the LLM better understand the AIOps inputs, establish task-input connections, and improve AIOps task prediction.

In this condition, we integrated retrieval-augmented generation into the two-stage LLM routing. By retrieving similar questions and their categories to the input question, RAG aids the general-purpose LLM in determining the input's task category. This led to the improvement of the CoE-Ops framework in the scenarios with high-level AIOps tasks.

Similar to other RAG approaches, the RAG process in our CoE-Ops can be divided into two sub-phases: Off-line and On-line, as abstractly shown in \eqref{eq:rag general}. 
% , and a knowledge base is constructed following the procedure outlined in \eqref{eq:context knowledge}
\begin{equation}
P(o|q)=\sum_{c\in\mathcal{C}}P(a|q,c)P(c|q),
\label{eq:rag general}
\end{equation}
where $q$ denotes the encoded vector of the query, $c$ represents the encoded vector of the context, and $o$ denotes the output of the LLM classifier.

During the Off-line stage, existing textual data is encoded, as shown in \eqref{eq:context encoder}. 
\begin{equation}
c = \text{Encoder}_{\text{RAG}}(C),
\label{eq:context encoder}
\end{equation}
where $C$ denotes the context data.
% V_{\text{context}} = \text{Encoder}_{\text{RAG}}(C)

% \begin{equation}
% \mathcal{K}=\{v_{c_i}|v_{c_i}=\mathrm{Encoder}_\mathrm{context}(c_i),c_i\in\mathcal{C}\}
% \label{eq:context knowledge}
% \end{equation}

In the On-line stage, the input AIOps query is first encoded into a vector by the encoder, as shown in \eqref{eq:query encoder}.

\begin{equation}
q = \text{Encoder}_{\text{RAG}}(Q),
\label{eq:query encoder}
\end{equation}
where $Q$ denotes the query data.
% V_{\text{query}} = \text{Encoder}_{\text{RAG}}(Q)

After obtaining the input AIOps query vector and knowledge base vectors, we perform retrieval to find the knowledge base vectors most similar to the input vector. The retrieval process is described by \eqref{eq:retriever}.

\begin{equation}
P(c|q)=\frac{\exp(\sin(q,c))}{\sum_{c\in\mathcal{C}}\exp(\sin(q,c))} .
\label{eq:retriever}
\end{equation}

The formula for the Retriever's similarity calculation is shown in \eqref{eq:simularity}.

\begin{equation}
\mathrm{sim}(q,c)=q \cdot c .
\label{eq:simularity}
\end{equation}

After incorporating the RAG technique, we retrieve similar problems to the input question, using them as context in the prompt. The improved prompt is shown in Prompt 3.
\begin{tcolorbox}[title = {Prompt 3 - Classifier with RAG}]
You are a classifier that can categorize questions into specific tasks. Your job is to analyze the following given question and determine which task from the provided list it most likely belongs to. \\
The tasks are as follows: \{$task~list$\}.\\
The question is: \\
"\{$question$\} \\ 
A.\{$option_A$\} \\ 
B.\{$option_B$\} \\ 
C.\{$option_C$\} \\ 
D.\{$option_D$\}".\\
You can refer to the following examples of questions and their corresponding tasks to decide the current question's task: \{$context$\}\\
Provide your answer in the format: "**Task: [$selected~task$] **".
\end{tcolorbox}

\section{Experiment}
\label{Experiment}

\subsection{Experimental Setup}
% To evaluate the performance of our developed CoDER in the field of software engineering, we conducted testing using the DevOps-Eval dataset, a comprehensive evaluation dataset specifically designed for large models in the DevOps domain. This repository contains a wealth of multiple-choice questions related to DevOps and AIOps. Based on language, it is divided into two sub-datasets: DevOps-Eval English and DevOps-Eval Chinese. The DevOps-Eval English dataset primarily focuses on AIOps. On the other hand, DevOps-Eval Chinese covers the entire DEVOPS process. The category of this dataset is presented in Tab.~\ref{tab:Dataset info of DEVOPS-EVAL}.

To validate the effectiveness of our designed CoE-Ops in the complex domain of AIOps Question-Answering, we evaluated its performance using the DevOps-Eval\footnote{https://hf-mirror.com/datasets/codefuse-ai/CodeFuse-DevOps-Eval} benchmark. DevOps-Eval is a comprehensive evaluation dataset specifically designed for large language models in the DevOps domain. This repository primarily contains a substantial collection of multiple-choice questions related to DevOps and AIOps, categorized into two subsets by language: DevOps-Eval English and DevOps-Eval Chinese. The DevOps-Eval English subset primarily covers low-level AIOps tasks, with its scope detailed in Table 1, while DevOps-Eval Chinese encompasses the comprehensive DevOps lifecycle, representing high-level AIOps tasks, as outlined in Tab.~\ref{tab:Dataset info of DEVOPS-EVAL}.

\begin{table}[H]
\caption{Dataset info of DevOps-EVAL}
\label{tab:Dataset info of DEVOPS-EVAL}
\begin{center}
\begin{tabular}{lc|lc}
\toprule
\multicolumn{2}{c}{\textbf{DEVOPS-EVAL English$^{\mathrm{a}}$}} & \multicolumn{2}{c}{\textbf{DEVOPS-EVAL Chinese$^{\mathrm{b}}$}} \\
\midrule
\textbf{Task} & \textbf{Sample} & \textbf{Task} & \textbf{Sample}\\
\midrule
LogParser & 350 & Build & 218\\
RootCauseAnalysis & 250  & Code & 1321\\
TimeSeriesAnomalyDetection & 300 & Deploy & 255\\
TimeSeriesClassification & 200 & Monitor & 216\\
TimeSeriesForecasting & 320 & Operate & 2041\\
 &  & Plan & 66\\
 &  & Release & 212 \\
 &  & Test & 228 \\
% \midrule
% Totals & 1420& Totals & 4557 \\
\bottomrule
\multicolumn{4}{l}{\footnotesize$^{\mathrm{a}}${Can be treated as "dataset with low-level tasks".}} \\
\multicolumn{4}{l}{\footnotesize$^{\mathrm{b}}${Can be treated as "dataset with high-level tasks".}} \\
\end{tabular}
\label{tab1}
\end{center}
\end{table}

\begin{table*}[htbp]
\caption{Task-Expert Mapping and Classifier Settings}
\label{tab:Task-Expert Mapping settings}
\begin{center}
\begin{tabular}{lcc|lcc}
\toprule
\textbf{Task Set A} & \textbf{Expert Set 1} & \textbf{Expert Set 2} & \textbf{Task Set B} & \textbf{Expert Set 3} & \textbf{Expert Set 4}\\
\midrule
Log Parser & Internlm-chat-7B\footnotemark[1] & Ministral-8b\footnotemark[2] & Build & Internlm-chat-7b & Gemma-2-27b-it\footnotemark[2]\\
Root Cause Analysis & CodeFuse-DevOps-Model-7B-Chat\footnotemark[1] & Ministral-8b & Code & Qwen2-7B-Instruct\footnotemark[1] & Doubao-1.5-lite-32k\footnotemark[3] \\
Time Series Anomaly Detection & CodeFuse-DevOps-Model-7B-Base\footnotemark[1] & Glm-4-flash\footnotemark[3] & Deploy & Internlm-chat-7b & Doubao-1.5-lite-32k\\ 
Time Series Classification & Internlm-7B\footnotemark[1] & Codegeex-4\footnotemark[3] & Monitor & Mathstral-7B-v0.1\footnotemark[1] & Gemma-2-27b-it\\
Time Series Forecasting & Internlm-chat-7B & Ministral-8b & Operate & Qwen2-7B-Instruct & Gemma-2-27b-it\\
 &  &  & Plan & Qwen2-7B-Instruct & Glm-4-flash\footnotemark[3]\\
 &  &  & Release & Mathstral-7B-v0.1 & Gemma-2-27b-it\\
 &  &  & Test & Qwen2-7B-Instruct & Doubao-1.5-lite-32k\\
\midrule
\textbf{Classifier 1} & \multicolumn{5}{c}{DeepSeek-R1-Distill-Qwen-7B\footnotemark[1]} \\
\midrule
\textbf{Classifier 2} & \multicolumn{5}{c}{DeepSeek-V3\footnotemark[2]} \\
\bottomrule
\multicolumn{6}{l}{\footnotesize[1]{Depolyed Locally}\label{fn:local}} \\
\multicolumn{6}{l}{\footnotesize[2]{Depolyed through API, base url: https://openrouter.ai/api/v1}\label{fn:openrouter}} \\
\multicolumn{6}{l}{\footnotesize[3]{Depolyed through API, base url: https://o3.fan/v1}\label{fn:o3.fan}} \\
% \multicolumn{2}{l}{\footnotesize $^{\mathrm{a}}$Sample of a Table footnote.}
\end{tabular}
\end{center}
\end{table*}

To evaluate the performance of numerous expert models across diverse task domains, we established a comprehensive benchmark and constructed the "Task-Expert" mapping presented in Tab.~\ref{tab:Task-Expert Mapping settings}, where Task Set A represents a low-level AIOps task and Task Set B constitutes a high-level AIOps task. For Set 1 and Set 3 in Tab.~\ref{tab:Task-Expert Mapping settings}, we deploy the corresponding expert models locally for inference due to their moderate parameter size. Regarding Set 2 and Set 4 in Tab.~\ref{tab:Task-Expert Mapping settings}, the substantial parameter scale of these expert models precludes local deployment. We directly invoke these models via API interfaces provided by open-source platforms for inference, since our proposed CoE-Ops framework requires neither fine-tuning nor training of the models.

Notably, to verify that CoE-Ops framework possesses good task and expert extensibility, when switching among the four sets, we only modified the prompts and the “task-expert” mapping, without altering the model architecture or retraining and fine-tuning the models.

For experimental evaluation metrics, we employed numerical indicators including accuracy, precision, recall, and F1-score for classification and question-answering tasks to quantify the results. Additionally, we visualized the experimental outcomes using confusion matrix heatmaps and model capability radar charts.

% We have analyzed the performance of numerous expert models in various task domains and established the 'Task-Expert' mapping presented in Tab.~\ref{tab:Task-Expert Mapping settings}. For Expert Set 1 and Expert Set 3 in Tab.~\ref{tab:Task-Expert Mapping settings}, given that these expert models have a moderate number of parameters, we opt to deploy and conduct inference of these expert models locally. As for Expert Set 2 and Expert Set 4 in Tab.~\ref{tab:Task-Expert Mapping settings}, due to the excessive number of parameters of these expert models, which makes local deployment infeasible, and considering that our proposed CoDER framework does not require fine-tuning or training of any models, we directly utilize the API interfaces provided by open-source platforms to call and perform inference of these models.

% \subsection{Research Questions}
% To evaluate the performance of the CoE-Ops, we have formulated the following three research questions (RQs):
% \begin{itemize}
% \item \textbf{RQ1 (Effectiveness):} Can CoE-Ops, using an ensemble learning paradigm, mitigate the competency gaps between different LLMs?

% \item \textbf{RQ2 (Scalability):} How can CoE-Ops be incorporated into ensemble learning for multitask learning in the AIOps domain?

% \item \textbf{RQ3 (Comparison):} Does the integration of smaller models via CoE-Ops enable performance that surpasses that of larger models?

% \end{itemize}

% ---------------------------------------------------------------------------------
\subsection{RQ1: CoE-Ops Effectiveness Evaluation}
\label{CoE-Ops Test Results}

To validate that our proposed CoE-Ops framework can balance capability disparities among models through ensemble learning across diverse model combinations, we applied CoE-Ops with different classifiers to expert collaborations (Expert Sets 1-4) on Task Set A and Task Set B from Tab.~\ref{tab:Task-Expert Mapping settings}. Specifically, we employed both the locally deployed DeepSeek-R1-Distill-Qwen-7B (Classifier 1) and the remotely accessed DeepSeek-V3 (Classifier 2) as task classifiers. For the RAG component, we employed the eval split from the DEVOPS-EVAL dataset as the context. The all-MiniLM-L6-v2 model was used as the encoder to encode both contexts and inputs into vector representations. Inputs were routed to corresponding AIOps experts within Expert Sets 1-4 based on the classification results.

\begin{figure}[htbp]
\centerline{\includegraphics[width=0.45\textwidth]{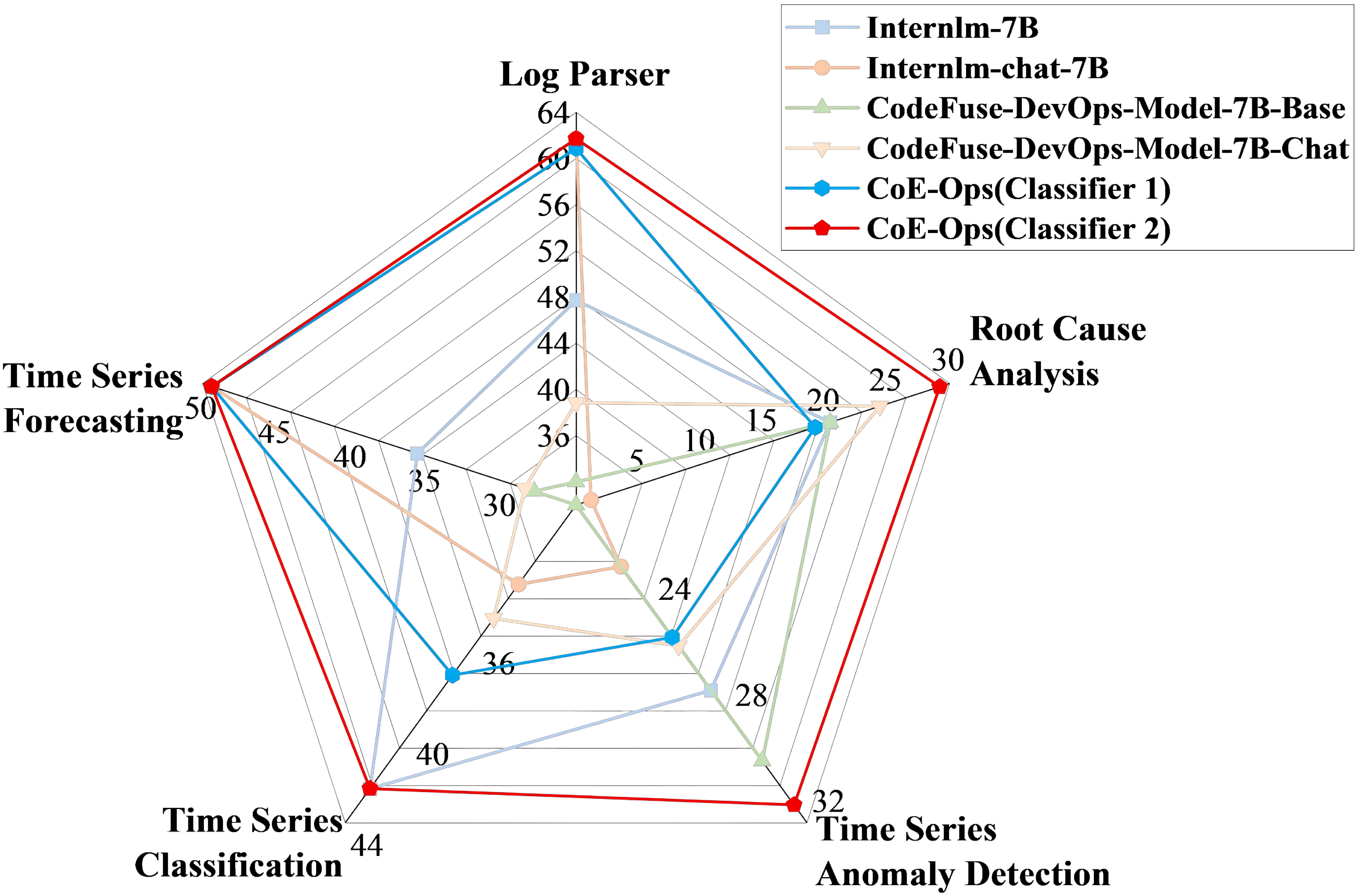}}
\caption{Capability Radar Chart of CoE-Ops with Expert Set 1 on DevOps-EVAL English (TASK SET A)}
\label{fig:ladar_en_small}
\end{figure}

\begin{figure}[htbp]
\centerline{\includegraphics[width=0.46\textwidth]{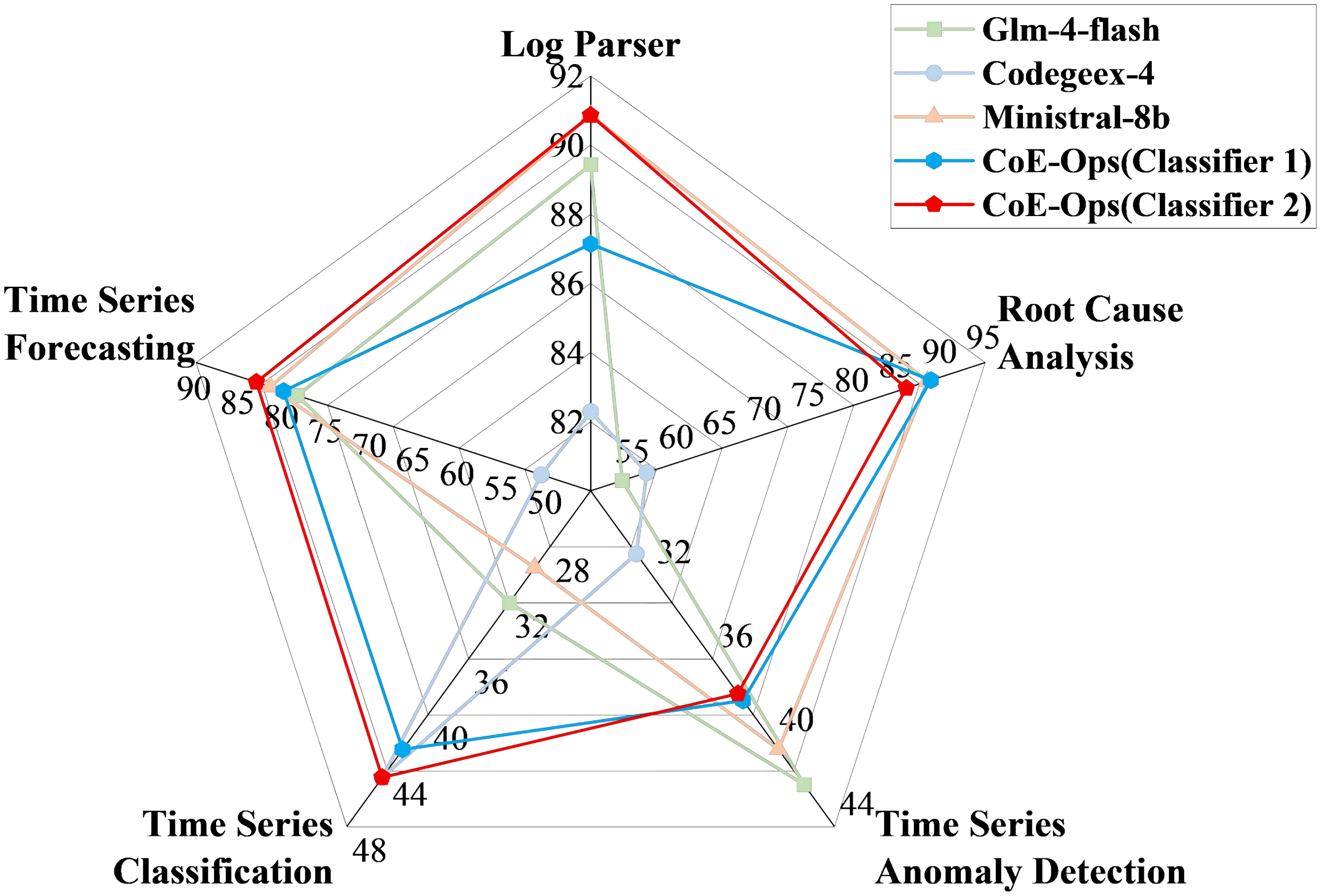}}
\caption{Capability Radar Chart of CoE-Ops with Expert Set 2 on DevOps-EVAL English (TASK SET A)}
\label{fig:ladar_en_large}
\end{figure}

\begin{figure}[htbp]
\centerline{\includegraphics[width=0.46\textwidth]{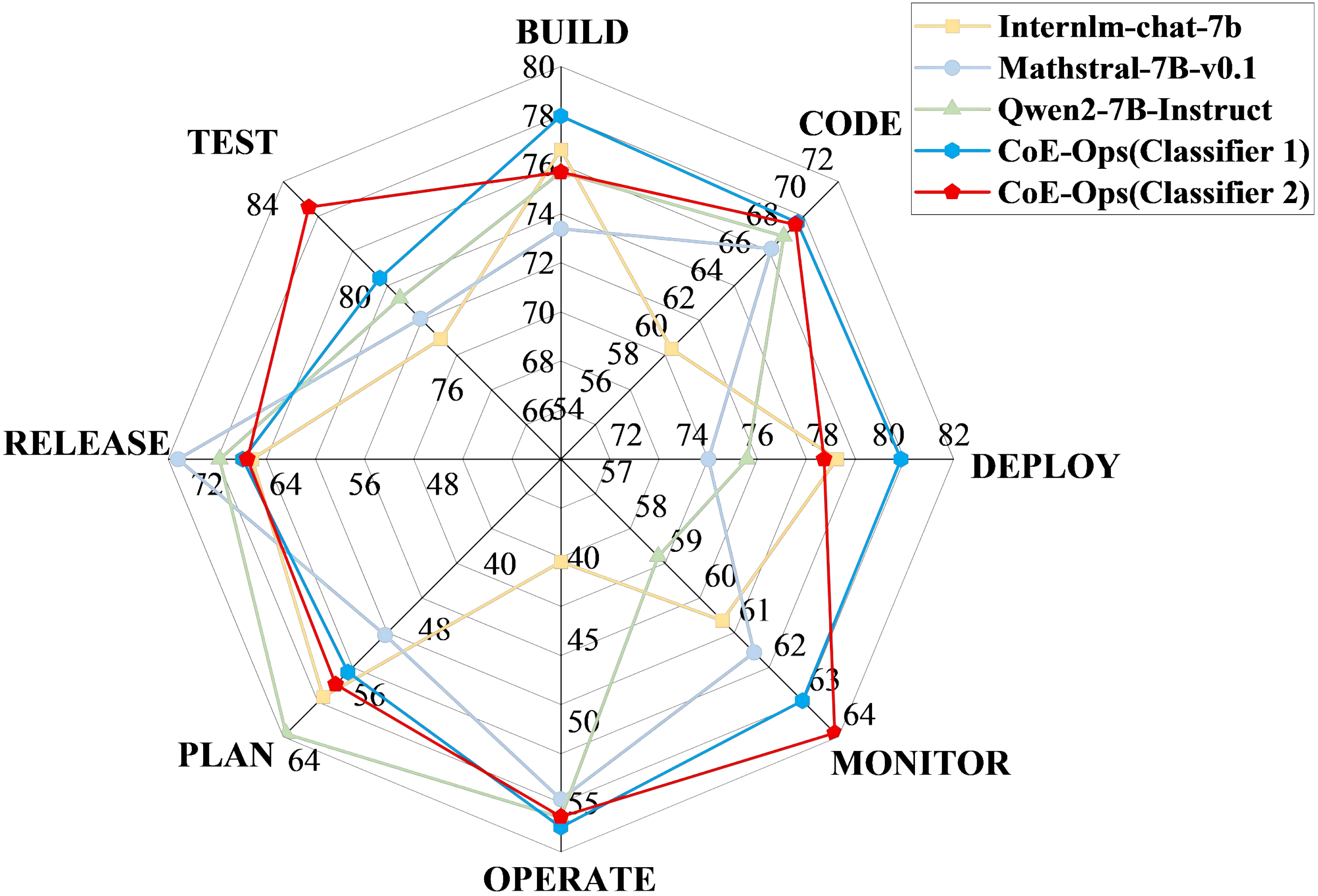}}
\caption{Capability Radar Chart of CoE-Ops with Expert Set 3 on DevOps-EVAL Chinese (TASK SET B)}
\label{fig:ladar_zh_small}
\end{figure}

\begin{figure}[htbp]
\centerline{\includegraphics[width=0.46\textwidth]{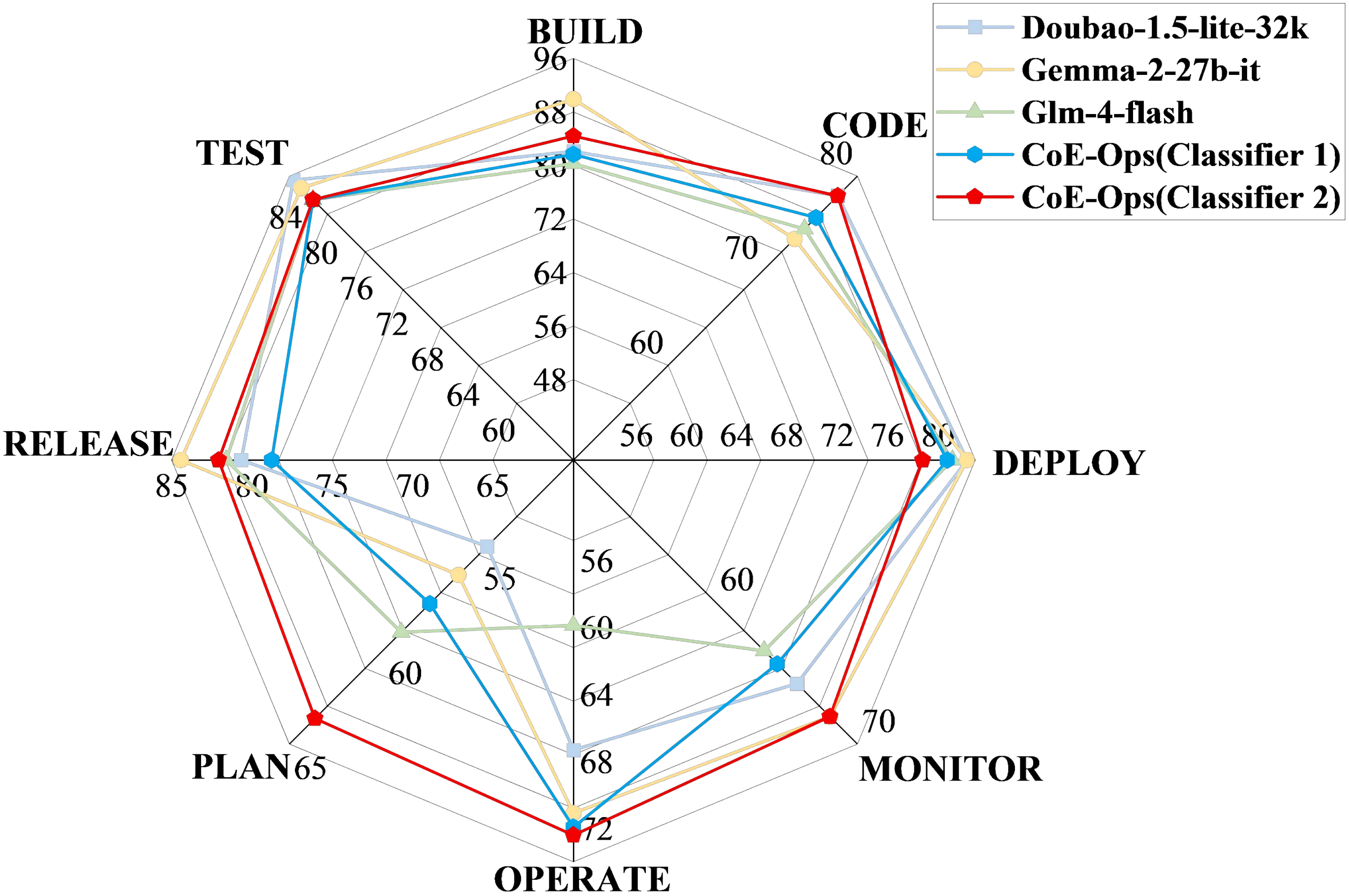}}
\caption{Capability Radar Chart of CoE-Ops with Expert Set 4 on DevOps-EVAL Chinese (TASK SET B)}
\label{fig:ladar_zh_large}
\end{figure}

We measured metrics such as answer accuracy for CoE-Ops and its utilized experts, and constructed capability radar charts for the models. The experimental results were subsequently organized and aggregated according to the Expert Sets. Specifically, results for Expert Set 1 are presented in Tab.~\ref{tab:DEVOPS_en small} and Fig.~\ref{fig:ladar_en_small}, Expert Set 2 in Tab.~\ref{tab:DEVOPS_en large} and Fig.~\ref{fig:ladar_en_large}, Expert Set 3 in Tab.~\ref{tab:DEVOPS_zh small} and Fig.~\ref{fig:ladar_zh_small}, and Expert Set 4 in Tab.~\ref{tab:DEVOPS_zh large} and Fig.~\ref{fig:ladar_zh_large}.

% After experimentally validating the efficacy of the CoOpERa task classifier, we proceeded to evaluate the performance of the complete CoOpERa framework using distinct expert sets across both the DevOps-Eval English and DevOps-Eval Chinese datasets, and summarized the results in tabular format. Specifically, Tab.~\ref{tab:DEVOPS_en small} presents the performance of CoOpERa on the DevOps-Eval English dataset using Expert Set 1, while Tab.~\ref{tab:DEVOPS_en large} illustrates its results on the same dataset with Expert Set 2. Similarly, Tab.~\ref{tab:DEVOPS_zh small} and Tab.~\ref{tab:DEVOPS_zh large} detail the outcomes for CoDER on the DevOps-Eval Chinese dataset when utilizing Expert Set 3 and Expert Set 4, respectively.

% ---------------------------------------------------------------------------------

\begin{table}[htbp]
\caption{Performance of CoE-Ops with Expert Set 1 on DEVOPS-EVAL English (Task Set A)}
\label{tab:DEVOPS_en small}
\begin{center}
\begin{tabular}{lcccc}
\toprule
\textbf{Models} & \textbf{Acc(\%)} & \textbf{Prec(\%)} & \textbf{Rec(\%)} & \textbf{F1(\%)} \\
\midrule
Internlm-7B & 35.07 & 37.05 & 35.07 & 34.36 \\
Internlm-chat-7B & 35.99 & 39.47 & 35.99 & 35.42 \\
CodeFuse-7B-Base$^{\mathrm{a}}$ & 28.17 & 29.57 & 28.17 & 25.39 \\
CodeFuse-7B-Chat$^{\mathrm{b}}$ & 30.56 & 31.71 & 30.56 & 30.36 \\
\midrule
% CoDER & 39.86 & 43 & 40 & 39 \\
CoE-Ops(Classifier 1) & 40.07 & 42.40 & 40.07 & 39.6 \\
\textbf{CoE-Ops(Classifier 2)} & \textbf{44.08} & \textbf{46.82} & \textbf{44.08} & \textbf{43.58} \\
\bottomrule
\multicolumn{5}{l}{\footnotesize $^{\mathrm{a}}$Model's full name: CodeFuse-DevOps-Model-7B-Base.} \\
\multicolumn{5}{l}{\footnotesize $^{\mathrm{b}}$Model's full name: CodeFuse-DevOps-Model-7B-Chat.} \\
\end{tabular}
\end{center}
\end{table}

As indicated in Tab.~\ref{tab:DEVOPS_en small}, the CoE-Ops framework employing two classifiers demonstrates significant improvements over individual AIOps expert models across metrics including Accuracy, Precision, Recall, and F1-score. Specifically, Accuracy shows respective improvements of 4\% and 8\% compared to the best-performing standalone AIOps model. The effectiveness of the CoE-Ops framework is further validated in Fig.~\ref{fig:ladar_en_small}.

% ---------------------------------------------------------------------------------

\begin{table}[htbp]
\caption{Performance of CoE-Ops with Expert Set 2 on DEVOPS-EVAL English (Task Set A)}
\label{tab:DEVOPS_en large}
\begin{center}
\begin{tabular}{lcccc}
\toprule
\textbf{Models} & \textbf{Acc(\%)} & \textbf{Prec(\%)} & \textbf{Rec(\%)} & \textbf{F1(\%)} \\
\midrule
Glm-4-flash & 62.54 & 64.50 & 62.54 & 63.16 \\
Codegeex-4 & 54.44 & 63.84 & 54.44 & 58.65 \\
Ministral-8b & 68.38 & 69.07 & 68.38 & 68.70 \\
% \midrule
% Mixtral-8x7b-instruct & 55.56 & 61.15 & 55.56 & 57.99 \\
% Random-CoE & 59.15 & 62.63 & 59.15 & 60.84 \\
% Bench-CoE & 68.94 & 70.30 & 68.94 & 69.58 \\
\midrule
% CoDER & 68.94 \\
CoE-Ops(Classifier 1) & 69.15 & 71.13 & 69.15 & 70.10 \\
\textbf{CoE-Ops(Classifier 2)} & \textbf{70.49} & \textbf{72.29} & \textbf{70.49} & \textbf{71.31} \\
\bottomrule
% \multicolumn{2}{l}{\footnotesize $^{\mathrm{a}}$Sample of a Table footnote.}
\end{tabular}
\end{center}
\end{table}

As shown in Tab.~\ref{tab:DEVOPS_en large}, the CoE-Ops framework utilizing two classifiers achieves balanced capability enhancement across varied expert configurations—both in quantity and type—on the same task. This demonstrates the scalability of our CoE-Ops framework with respect to model composition, as further evidenced in Fig.~\ref{fig:ladar_en_small} and Fig.~\ref{fig:ladar_en_large}.

% ---------------------------------------------------------------------------------

\begin{table}[htbp]
\caption{Performance of CoE-Ops with Expert Set 3 on DEVOPS-EVAL Chinese (Task Set B)}
\label{tab:DEVOPS_zh small}
\begin{center}
\begin{tabular}{lcccc}
\toprule
\textbf{Models} & \textbf{Acc(\%)} & \textbf{Prec(\%)} & \textbf{Rec(\%)} & \textbf{F1(\%)} \\
\midrule
Internlm-chat-7b & 54.2 & 53.63 & 54.20 & 53.56\\
Mathstral-7B-v0.1 & 62.74 & 62.77 & 62.74 & 62.47\\
Qwen2-7B-Instruct & 63.57 & 64.44 & 63.57 & 63.32\\
\midrule
\textbf{CoE-Ops(Classifier 1)} & \textbf{64.52} & \textbf{64.93} & \textbf{64.52} & \textbf{64.24} \\
CoE-Ops(Classifier 2) & 64.14 & 64.44 & 64.14 & 63.86 \\
\bottomrule
% \multicolumn{2}{l}{\footnotesize $^{\mathrm{a}}$Sample of a Table footnote.}
\end{tabular}
\end{center}
\end{table}

For high-level AIOps tasks such as Task Set B, despite their increased task classification difficulty, our CoE-Ops framework consistently outperforms individual AIOps expert models. This capability enhancement is evidenced by the analysis presented in Tab.~\ref{tab:DEVOPS_zh small} and Fig.~\ref{fig:ladar_zh_small}.

% ---------------------------------------------------------------------------------

\begin{table}[htbp]
\caption{Performance of CoE-Ops with Expert Set 4 on DEVOPS-EVAL Chinese (Task Set B)}
\label{tab:DEVOPS_zh large}
\begin{center}
\begin{tabular}{lcccc}
\toprule
\textbf{Models} & \textbf{Acc(\%)}  & \textbf{Prec(\%)} & \textbf{Rec(\%)} & \textbf{F1(\%)} \\
\midrule
Doubao-1.5-lite-32k & 73.21 & 73.73 & 73.21 & 73.47 \\
Gemma-2-27b-it & 74.22 & 74.13 & 74.22 & 74.14 \\
Glm-4-flash & 68.60 & 68.23 & 68.6 & 68.26 \\
% \midrule
% Mixtral-8x7b-instruct & 65.26 & 66.89 & 65.26 & 65.94 \\
\midrule
CoE-Ops(Classifier 1) & 74.28 & 74.79 & 74.28 & 74.52 \\
\textbf{CoE-Ops(Classifier 2)} & \textbf{75.60} & \textbf{75.91} & \textbf{75.60} & \textbf{75.75} \\
\bottomrule
% \multicolumn{2}{l}{\footnotesize $^{\mathrm{a}}$Sample of a Table footnote.}
\end{tabular}
\end{center}
\end{table}

% Analysis of Tab.~\ref{tab:DEVOPS_en large} and Tab.~\ref{tab:DEVOPS_zh large} demonstrates that our CoDER achieves measurable performance improvements even when applied to online-deployed expert models accessed via API calls.
Similarly, by synthesizing results from Tab.~\ref{tab:DEVOPS_zh small} and Tab.~\ref{tab:DEVOPS_zh large}, we observe that our CoE-Ops framework also exhibits model scalability on high-level AIOps tasks, it consistently enhances overall accuracy across model combinations involving both locally and remotely deployed models. This capability is further demonstrated in Fig.~\ref{fig:ladar_zh_small} and Fig.~\ref{fig:ladar_zh_large}.

% ---------------------------------------------------------------------------------

% To visually demonstrate the competency profiles of CoDER and its sub-experts across various task domains, we generated capability radar charts. Specifically, Fig.~\ref{fig:ladar_en_small} presents the performance of CoDER on the DevOps-Eval English dataset using Expert Set 1, while Fig.~\ref{fig:ladar_en_large} illustrates its results on the same dataset with Expert Set 2. Similarly, Fig.~\ref{fig:ladar_zh_small} and Fig.~\ref{fig:ladar_zh_large} depict the outcomes for CoDER on the DevOps-Eval Chinese dataset when employing Expert Set 3 and Expert Set 4, respectively.

In summary, through comprehensive analysis of Accuracy, Recall, F1-Score, and model capability radar charts across diverse expert configurations on multiple AIOps tasks, we demonstrate the effectiveness of the CoE-Ops framework in balancing heterogeneous model capabilities while establishing its scalability across varying model compositions.

\begin{tcolorbox}
\textbf{Answer to RQ1:} Experimental results demonstrate that our proposed CoE-Ops framework effectively balances capability discrepancies among diverse models across various tasks and expert settings. This integration ultimately achieves an overall performance improvement of up to approximately 8\%, confirming the effectiveness of our approach.
\end{tcolorbox}

% ---------------------------------------------------------------------------------

\subsection{RQ2: Classifier Scalability Validation}

Following the validation that our proposed CoE-Ops framework effectively balances capability disparities across different AIOps models, we conducted an ablation study on its core component, the Classifier, to assess its scalability for complex tasks in the AIOps domain. We evaluated two Classifiers employed by CoE-Ops (Classifier 1 and Classifier 2) on Task Set A and Task Set B, as detailed in Tab.~\ref{tab:Task-Expert Mapping settings}. Additionally, we tested the classification performance of a baseline Classifier without Retrieval-Augmented Generation enhancement. Task Set A and Task Set B differ in both the number of tasks and their hierarchical complexity. Testing on these two tasks thus allows coverage of the two Task Scalability Scenarios outlined in Section \ref{sec:Basic Concepts}.

We also evaluated the performance of the Bench-CoE framework, which utilizes a fine-tuned classifier, on both Task Set A and Task Set B in AIOps as a control. The general experimental results are presented in Tab.~\ref{tab:CoE-Ops router eng} (for Task Set A) and Tab.~\ref{tab:CoE-Ops router zh} (for Task Set B).

% Since we replaced the original two-stage expert router's task classifier with a non-fine-tuned general large model, we first tested its classification effectiveness. We used DeepSeek-R1-Distill-Qwen-7B as the new task classifier and optimized the input with task-classification-specific prompts from Subsections \ref{subsec: General Large Language Model} and \ref{subsec: Large Language Model Enhance with RAG}. For comparison, we calculated the random classification accuracy and tested the Bench CoE task classifier's accuracy on these tasks.

% We evaluated the CoDER task classifier on the DevOps-Eval English and DevOps-Eval Chinese datasets, with the results presented in Tab.~\ref{tab:Coopera router eng} and Tab.~\ref{tab:Coopera router zh} .

\begin{table}[htbp]
\caption{Classify Performance on DevOps-Eval English (Task Set A)}
\label{tab:CoE-Ops router eng}
\begin{center}
\begin{tabular}{lcccc}
\toprule
\textbf{Classifiers} & \textbf{Acc(\%)} & \textbf{Prec(\%)} & \textbf{Rec(\%)} & \textbf{F1(\%)}\\
\midrule
Random Select & 20.00 & - & - & - \\
Bench-CoE & 62.46 & 52.69 & 62.46 & 55.35 \\
\midrule
Classifier 1 w/o RAG & 77.11 & 87.66 & 77.11 & 81.52 \\
\textbf{Classifier 1} & \textbf{80.92}& \textbf{95.62} & \textbf{80.92} & \textbf{87.51} \\
\midrule
Classifier 2 w/o RAG & 100 & 100 & 100 & 100 \\
\textbf{Classifier 2} & \textbf{100}& \textbf{100} & \textbf{100} & \textbf{100} \\
\bottomrule
% \multicolumn{2}{l}{\footnotesize $^{\mathrm{a}}$Sample of a Table footnote.}
\end{tabular}
\end{center}
\end{table}
% CoOpERa(Classifier 2 w/o RAG)

\begin{table}[htbp]
\caption{Classify Performance on DevOps-Eval Chinese (Task Set B)}
\label{tab:CoE-Ops router zh}
\begin{center}
\begin{tabular}{lcccc}
\toprule
\textbf{Classifiers} & \textbf{Acc(\%)} & \textbf{Prec(\%)} & \textbf{Rec(\%)} & \textbf{F1(\%)}\\
\midrule
Random Select & 12.5 & - & - & - \\
Bench-CoE & 4.94 & 11.86 & 4.94 & 0.83 \\
\midrule
Classifier 1 w/o RAG & 13.91 & 32.65 & 13.91 & 14.66 \\
\textbf{Classifier 1} & \textbf{43.84} & \textbf{71.47} & \textbf{43.84} & \textbf{50.43} \\
\midrule
Classifier 2 w/o RAG & 24.93 & 41.67 & 24.93 & 26.54 \\
\textbf{Classifier 2} & \textbf{77.22} & \textbf{79.79} & \textbf{77.22} & \textbf{77.22} \\
\bottomrule
% \multicolumn{2}{l}{\footnotesize $^{\mathrm{a}}$Sample of a Table footnote.} $^*$
\end{tabular}
\end{center}
\end{table}
% ---------------------------------------------------------------------------------

\begin{figure}[htbp]
\centerline{\includegraphics[width=0.4\textwidth]{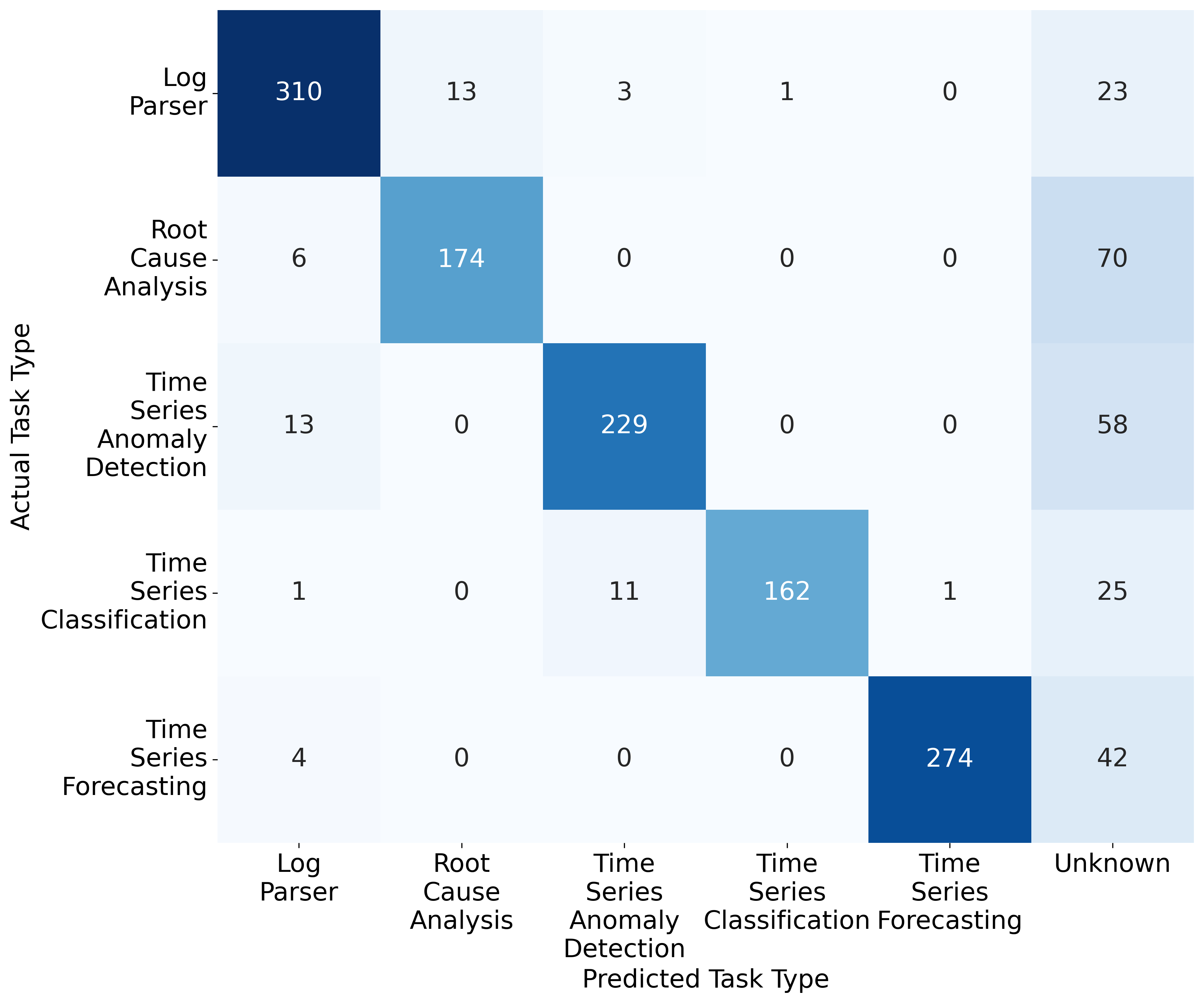}}
\caption{Heatmap Visualization of Classifier 1's Confusion Matrix on DevOps-EVAL English (Task Set A)}
\label{fig:router_en_rag}
\end{figure}

\begin{figure}[htbp]
\centerline{\includegraphics[width=0.4\textwidth]{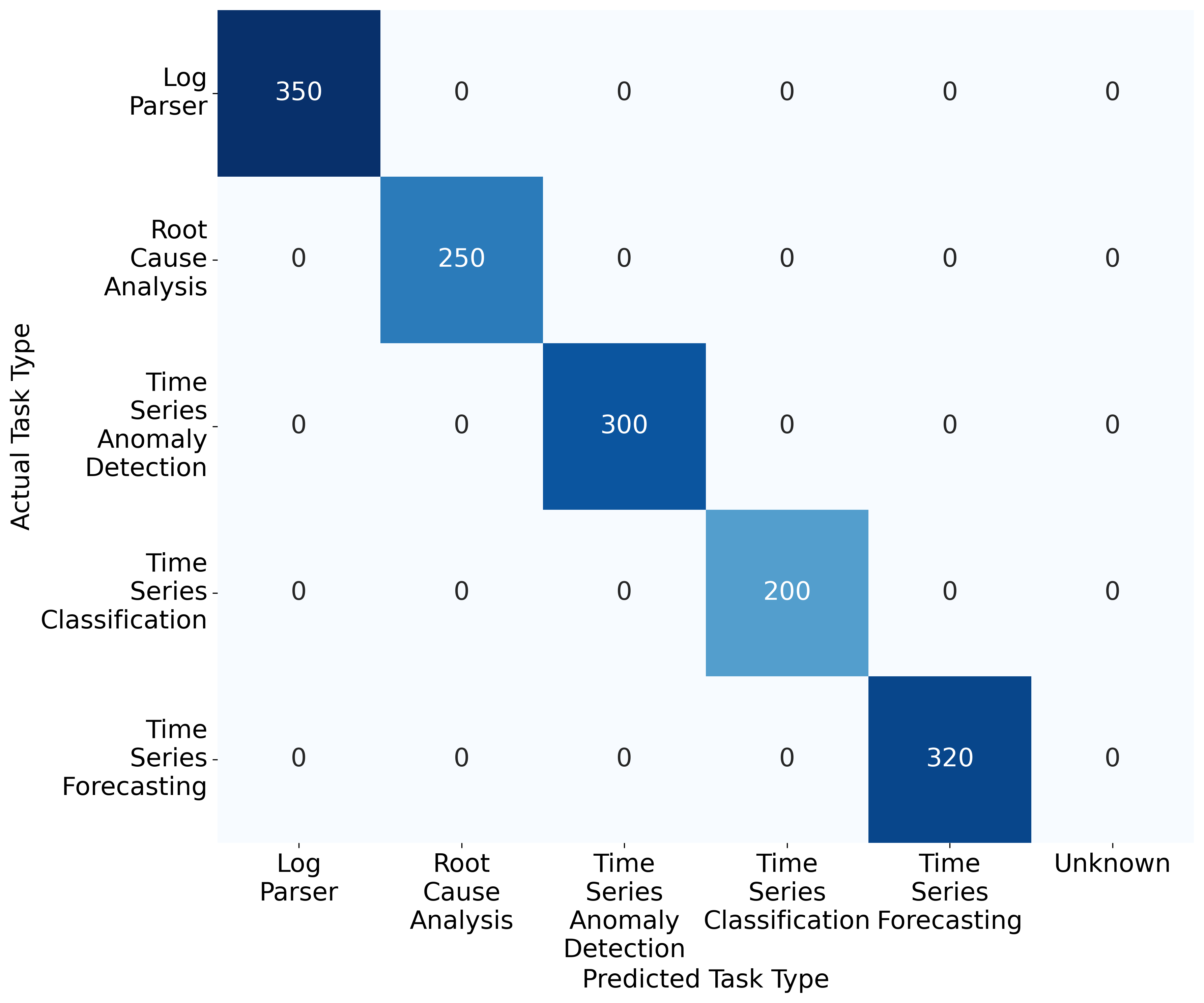}}
\caption{Heatmap Visualization of Classifier 2's Confusion Matrix on DevOps-EVAL English (Task Set A)}
\label{fig:router_en_rag_api}
\end{figure}

\begin{figure}[htbp]
\centerline{\includegraphics[width=0.4\textwidth]{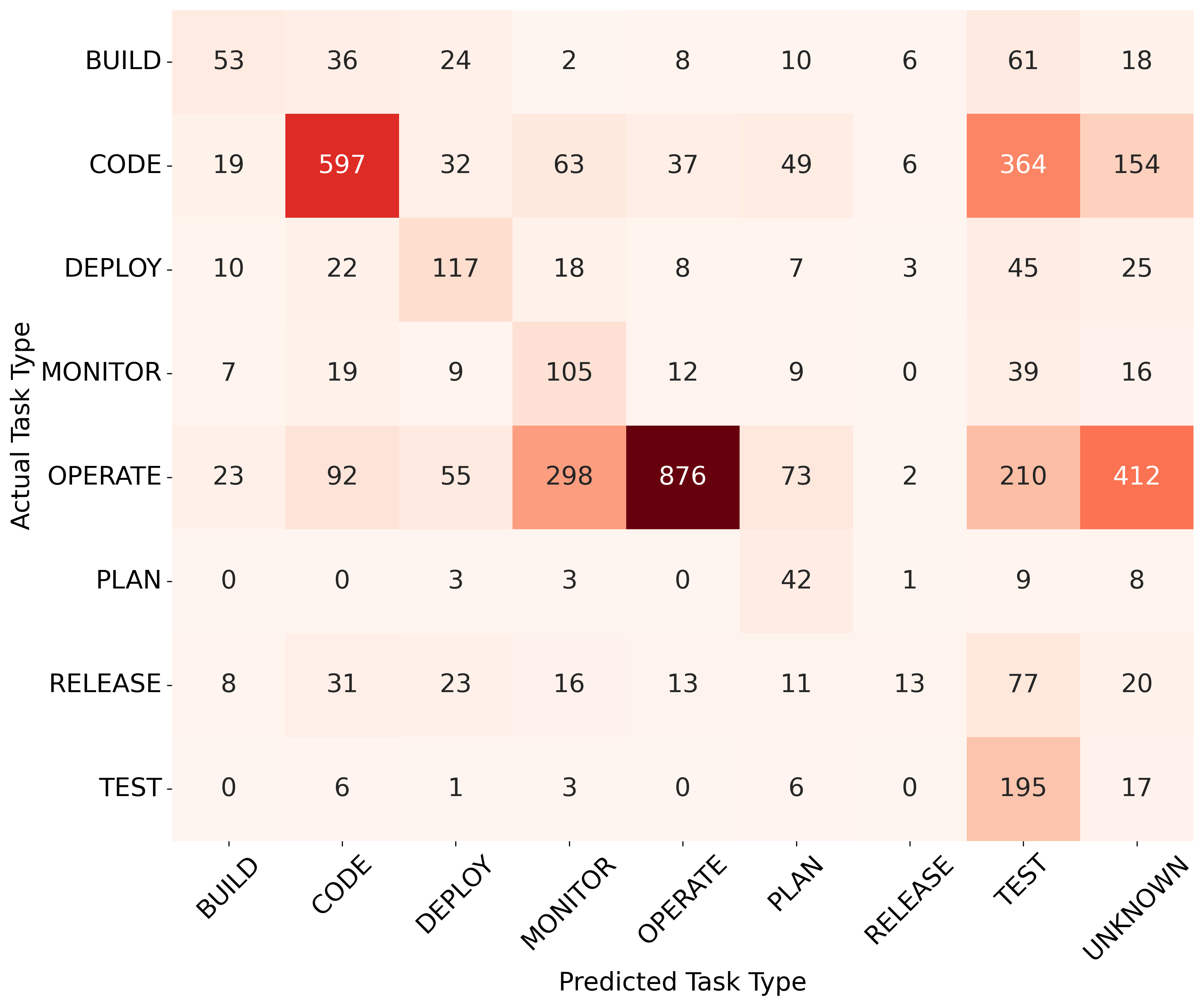}}
\caption{Heatmap Visualization of Classifier 1's Confusion Matrix on DevOps-EVAL Chinese (Task Set B)}
\label{fig:router_zh_rag}
\end{figure}

\begin{figure}[htbp]
\centerline{\includegraphics[width=0.4\textwidth]{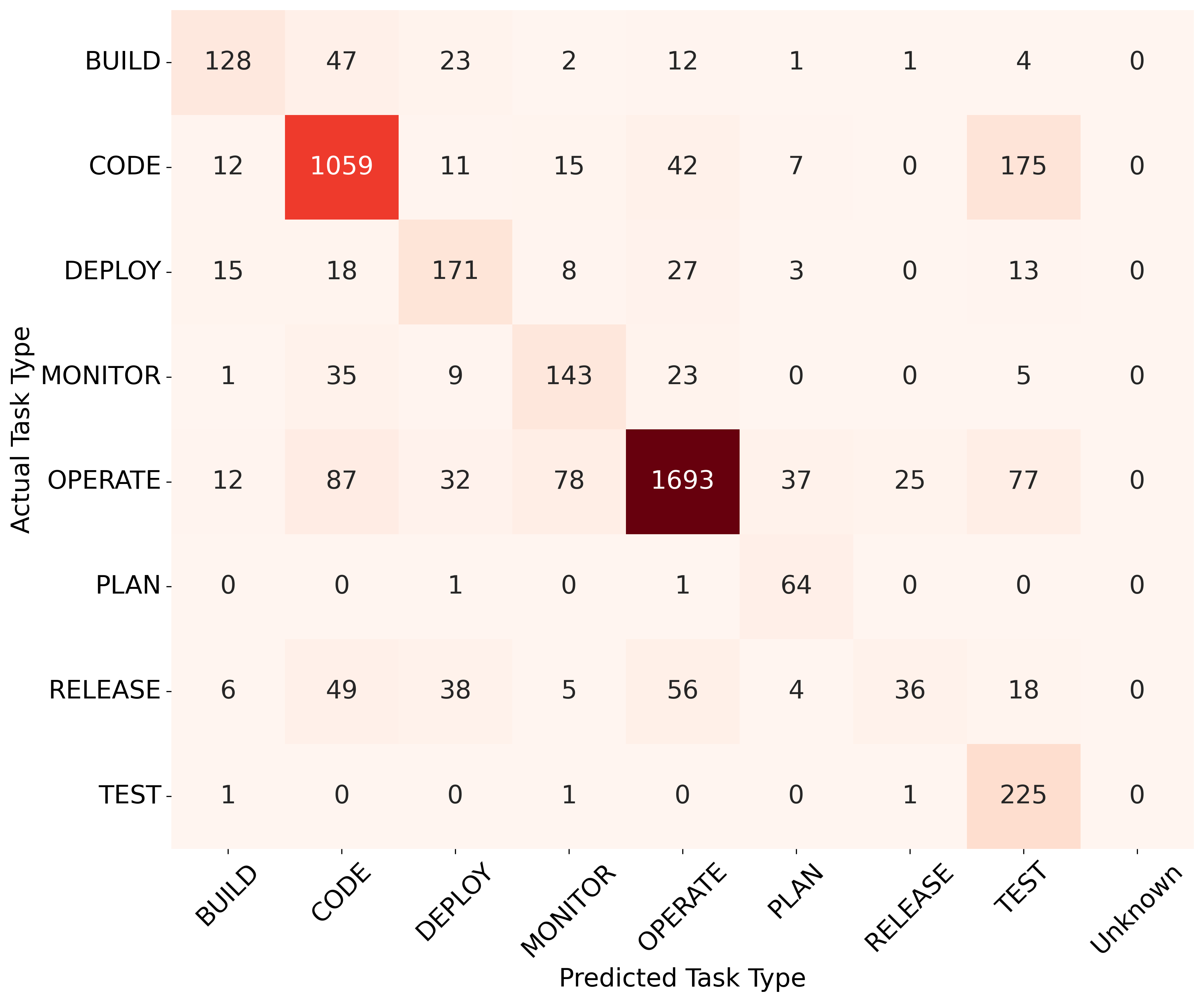}}
\caption{Heatmap Visualization of Classifier 2's Confusion Matrix on DevOps-EVAL Chinese (Task Set B)}
\label{fig:router_zh_rag_api}
\end{figure}

Furthermore, to facilitate a more intuitive analysis of the classification performance of the two Classifiers employed by the CoE-Ops framework on individual tasks within Task Set A and Task Set B, we visualized their results using heatmaps. The classification results for Task Set A are presented in Fig.~\ref{fig:router_en_rag} (Classifier 1) and Fig.~\ref{fig:router_en_rag_api} (Classifier 2), respectively. Similarly, the results for Task Set B are shown in Fig.~\ref{fig:router_zh_rag} (Classifier 1) and Fig.~\ref{fig:router_zh_rag_api} (Classifier 2).

% To better illustrate the results, we use confusion matrices in Fig.~\ref{fig:router_en_rag}, Fig.~\ref{fig:router_en_rag_api}, Fig.~\ref{fig:router_zh_rag} and Fig.~\ref{fig:router_zh_rag_api} to show the CoDER task classifier's performance across different task domains.

% \begin{figure}[htbp]
% \centerline{\includegraphics[width=0.4\textwidth]{figs_result/confusion/confusion_matrix_en.png}}
% \caption{Heatmap Visualization of CODER router's Confusion Matrix on DevOps-EVAL English}
% \label{fig:router_en}
% \end{figure}

Analysis of Tab.~\ref{tab:CoE-Ops router eng} reveals that Classifier 1 and Classifier 2, implemented without fine-tuning or retraining, achieved strong classification performance in Task Set A. Their classification accuracy surpassed that of the Bench-CoE framework, which uses a fine-tuned classifier. In particular, Classifier 2 achieved the classification accuracy 100\%, demonstrating its robust generalization capability. Furthermore, the classification accuracy of Classifier 1 showed a significant improvement after RAG integration. The heatmaps presented in Fig.~\ref{fig:router_en_rag} and Fig.~\ref{fig:router_en_rag_api} further validate the performance of both classifiers.
% , as indicated in Tab.~\ref{tab:CoE-Ops router eng}
% From Tab.~\ref{tab:Coopera router eng} and Fig.~\ref{fig:router_en_rag} and Fig.~\ref{fig:router_en_rag_api}, we can see that the CoDER task classifier, when enhanced with RAG technology, shows higher classification accuracy. It also tends to classify ambiguous questions as “Unknown” instead of forcing a classification, which reduces the risk of hallucinations in the classifier.

% \begin{figure}[htbp]
% \centerline{\includegraphics[width=0.4\textwidth]{figs_result/confusion/confusion_matrix_zh.png}}
% \caption{Heatmap Visualization of CODER router's Confusion Matrix on DevOps-EVAL Chinese}
% \label{fig:router_zh}
% \end{figure}

A comparison of Tab.~\ref{tab:CoE-Ops router eng} and Tab.~\ref{tab:CoE-Ops router zh} reveals that while the Bench-CoE framework, based on a fine-tuned classifier, demonstrates acceptable classification performance in the low-level AIOps task (Task Set A), its accuracy exhibits a marked degradation when the AIOps task scenario shifts to the high-level AIOps task (Task Set B). In contrast, although the performance of both Classifiers within our CoE-Ops framework also declined, their classification accuracy showed significant recovery, particularly for Classifier 2, upon augmentation with Retrieval-Augmented Generation technology. This robustly demonstrates the task scalability of our CoE-Ops framework within the complex AIOps task domain. This identical conclusion is further corroborated by the graphical evidence presented in Fig.~\ref{fig:router_zh_rag} and Fig.~\ref{fig:router_zh_rag_api}.

% By analyzing Tab.~\ref{tab:Coopera router zh}, Fig.~\ref{fig:router_zh_rag}, and Fig.~\ref{fig:router_zh_rag_api}, it is evident that in the DevOps-Eval Chinese dataset, the classification labels (e.g., DEPLOY, RELEASE) are more abstract compared to specific task types (e.g., Log Parser). This abstraction leads to ambiguity in distinguishing certain questions across multiple task domains (e.g., OPERATE and MONITOR), significantly increasing the difficulty of task classification. Consequently, when replacing the classifier solely with a general-purpose large language model (LLM) without leveraging Retrieval-Augmented Generation (RAG) techniques, the improvement in classification accuracy remains marginal. However, after integrating RAG-enhanced methods, CoDER's task classifier demonstrates a marked increase in classification accuracy, enabling effective task categorization and expert routing.

\begin{tcolorbox}
\textbf{Answer to RQ2:} We design CoE-Ops, which employs a general-purpose large language model as the task classifier. This classifier is enhanced using prompting and Retrieval-Augmented Generation (RAG) techniques to adapt to the complex task scenarios in AIOps. We conduct classification experiments on both low-level and high-level tasks. The experimental results demonstrate that our CoE-Ops achieves significantly higher task classification accuracy compared to other ensemble learning methods in AIOps, showing improvements of 37.54\% and 72.28\%, respectively.
\end{tcolorbox}

% ---------------------------------------------------------------------------------

\subsection{RQ3: Efficiency Validation}

\begin{figure}[htbp]
\centerline{\includegraphics[width=0.42\textwidth]{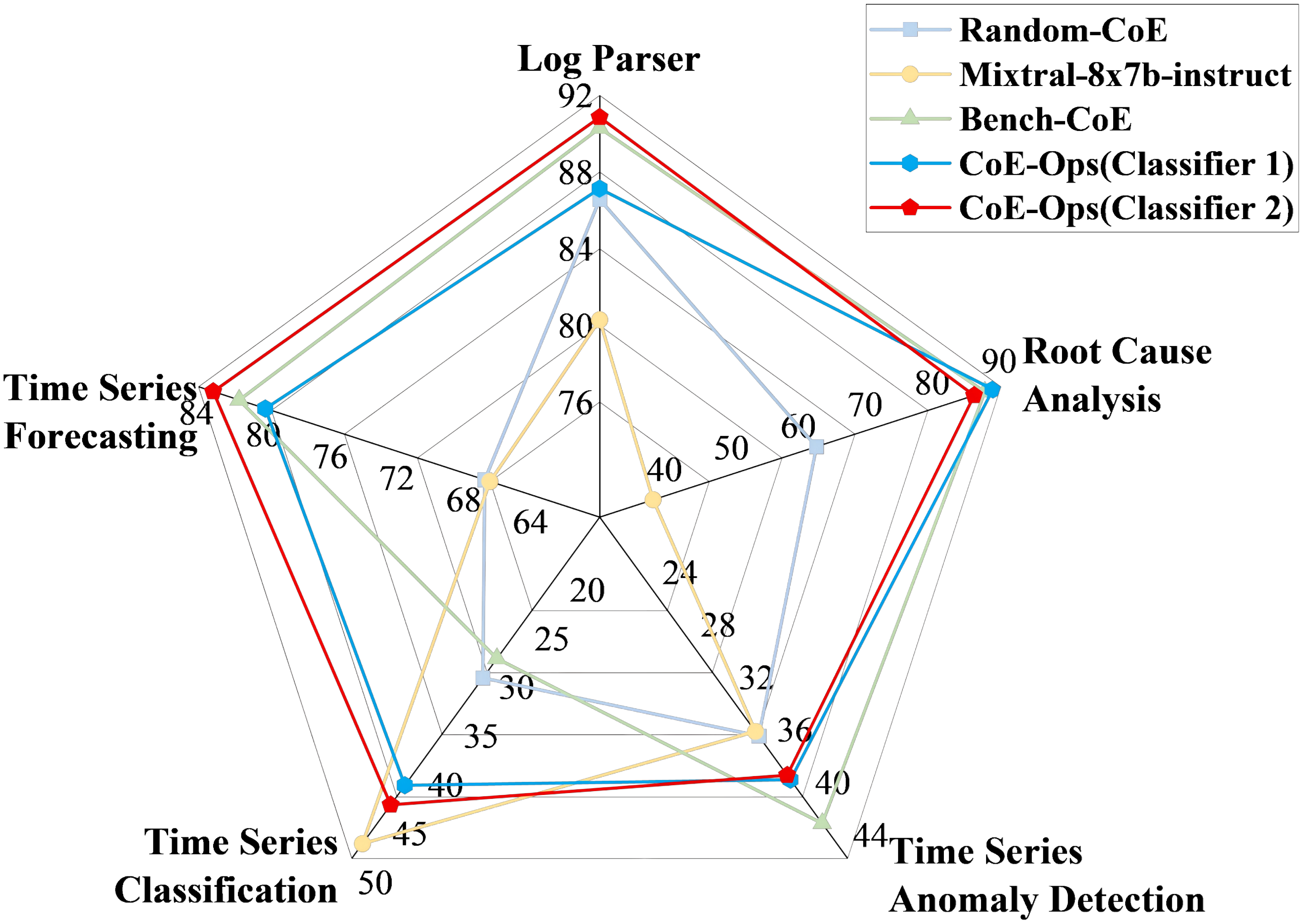}}
\caption{Capability Radar Chart of Comparative Experiments on DevOps-EVAL English (Task Set A)}
\label{fig:ladar_en_moe}
\end{figure}

\begin{figure}[htbp]
\centerline{\includegraphics[width=0.42\textwidth]{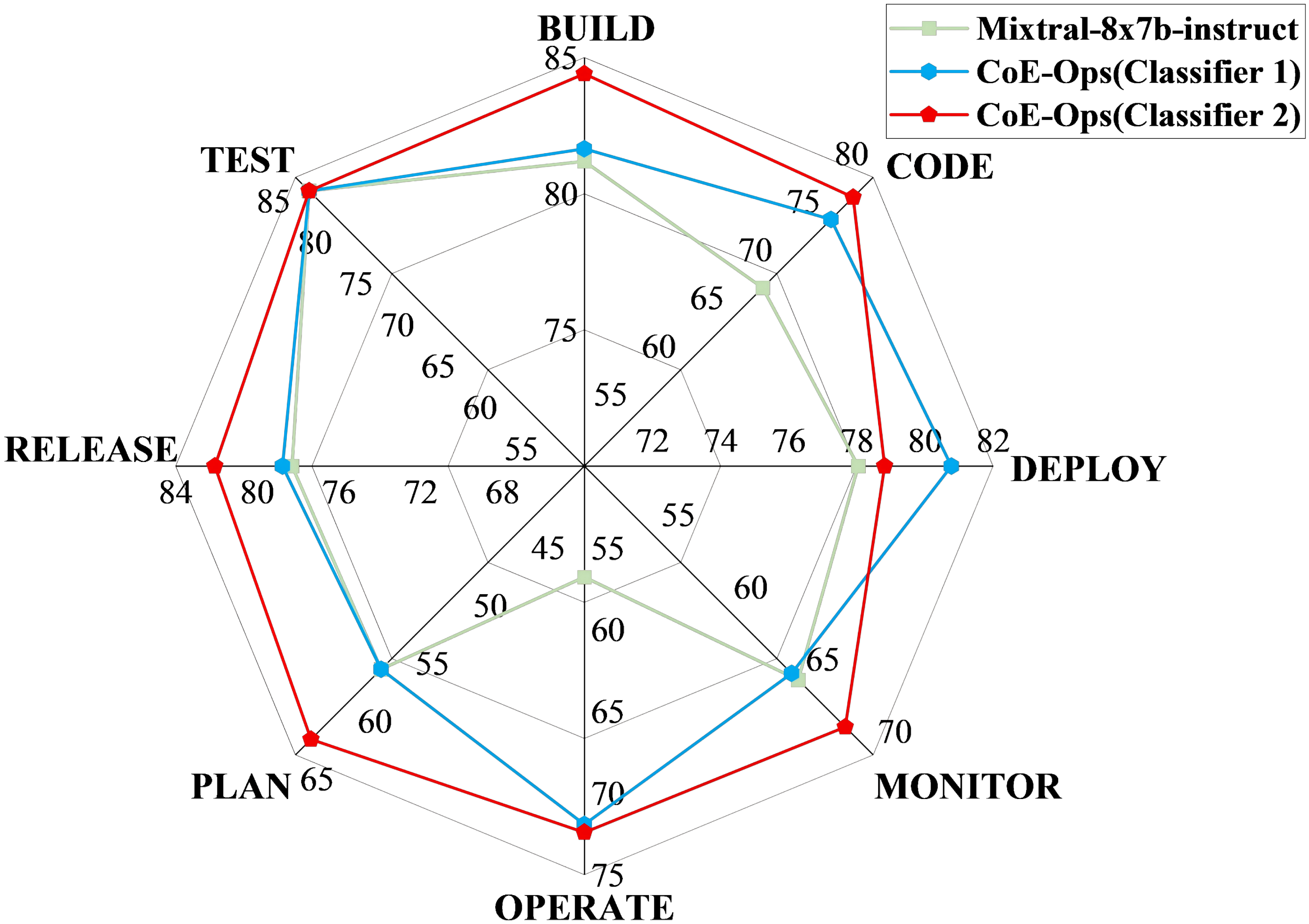}}
\caption{Capability Radar Chart of Comparative Experiments on DevOps-EVAL Chinese (Task Set B)}
\label{fig:ladar_zh_moe}
\end{figure}

Following the validation of CoE-Ops' effectiveness in balancing model capabilities and its classifier's task scalability, we further compared CoE-Ops against other CoE and MoE models. Notably, the total parameter count of the mixtral-8x7b-instruct model reached approximately 56B, while the largest model deployed by our CoE-Ops utilized 27B parameters. We evaluated these models separately on Task Set A and Task Set B. Bench-CoE and Random-CoE (CoE with entirely random model routing) were tested on Task Set A as control groups, while Bench-CoE was not tested as a control group on Task Set B due to its poor classification performance. The experimental results are presented in Tab.~\ref{tab:DEVOPS_en moe} and Tab.~\ref{tab:DEVOPS_zh moe}, respectively, and are also visualized in the model capability radar charts shown in Fig.~\ref{fig:ladar_en_moe} and Fig.~\ref{fig:ladar_zh_moe}.

% We evaluated these models separately on Task Set A and Task Set B. Additionally, Bench-CoE and Random-CoE (CoE with entirely random model routing) were tested on Task Set A as control groups.

\begin{table}[htbp]
\caption{Performance of CoE and MoE with Expert Set 2 on DEVOPS-EVAL English (Task Set A)}
\label{tab:DEVOPS_en moe}
\begin{center}
\begin{tabular}{lcccc}
\toprule
\textbf{Models} & \textbf{Acc(\%)} & \textbf{Prec(\%)} & \textbf{Rec(\%)} & \textbf{F1(\%)} \\
\midrule
Mixtral-8x7b-instruct & 55.56 & 61.15 & 55.56 & 57.99 \\
Random-CoE & 59.15 & 62.63 & 59.15 & 60.84 \\
Bench-CoE & 68.94 & 70.30 & 68.94 & 69.58 \\
\midrule
CoE-Ops(Classifier 1) & 69.15 & 71.13 & 69.15 & 70.10 \\
\textbf{CoE-Ops(Classifier 2)} & \textbf{70.49} & \textbf{72.29} & \textbf{70.49} & \textbf{71.31} \\
\bottomrule
\end{tabular}
\end{center}
\end{table}

\begin{table}[htbp]
\caption{Performance of CoE and MoE with Expert Set 4 on DEVOPS-EVAL Chinese (Task Set B)}
\label{tab:DEVOPS_zh moe}
\begin{center}
\begin{tabular}{lcccc}
\toprule
\textbf{Models} & \textbf{Acc(\%)}  & \textbf{Prec(\%)} & \textbf{Rec(\%)} & \textbf{F1(\%)} \\
\midrule
Mixtral-8x7b-instruct & 65.26 & 66.89 & 65.26 & 65.94 \\
\midrule
CoE-Ops(Classifier 1) & 74.28 & 74.79 & 74.28 & 74.52 \\
\textbf{CoE-Ops(Classifier 2)} & \textbf{75.60} & \textbf{75.91} & \textbf{75.60} & \textbf{75.75} \\
\bottomrule
% \multicolumn{2}{l}{\footnotesize $^{\mathrm{a}}$Sample of a Table footnote.}
\end{tabular}
\end{center}
\end{table}

% Additionally, building on our existing experiments, we incorporated ablation and controlled experiments. Specifically, we expanded the DevOps-Eval English dataset to include test results from random routing for expert selection and Bench CoE-based expert selection. Furthermore, we conducted comparative analyses against the mainstream MoE model, mixtral-8x7b-instruct, across both the DevOps-Eval English and DevOps-Eval Chinese datasets. The experimental results are presented in Tab.~\ref{tab:DEVOPS_en large} and Tab.~\ref{tab:DEVOPS_zh large}, while the corresponding capability radar charts are depicted in Fig.~\ref{fig:ladar_en_moe} and Fig.~\ref{fig:ladar_zh_moe}.

As indicated in Tab.~\ref{tab:DEVOPS_en moe}, CoE-Ops demonstrates superior overall capability in the complex domain of AIOps compared to existing CoE and MoE models. Analysis combining Tab.~\ref{tab:DEVOPS_en moe} and Tab.~\ref{tab:DEVOPS_zh moe} reveals that CoE-Ops, leveraging an ensemble of smaller models, comprehensively surpasses large models such as mixtral-8x7b-instruct in terms of overall performance. This conclusion is further supported by the evidence presented in Fig.~\ref{fig:ladar_en_moe} and Fig.~\ref{fig:ladar_zh_moe}.

% Analysis of the aforementioned charts reveals that CoOpERa achieves substantial improvements in DevOps capabilities compared to random routing and existing CoE and MoE models, thereby establishing its state-of-the-art (SOTA) performance.

\begin{tcolorbox}
\textbf{Answer to RQ3:} We conduct experiments comparing the performance of CoE-Ops integrated with small models against existing large models implemented via the MoE paradigm on the DEVOPS-EVAL dataset. The results experimentally demonstrate that, when appropriate small models are selected, CoE-Ops's integration of these small models achieves performance surpassing that of large models.
\end{tcolorbox}

% ---------------------------------------------------------------------------------

\section{Threats to Validity}

We acknowledge the following potential threats to the validity of our study and discuss our mitigation strategies:

\paragraph{Internal validity} Internal threats primarily center on the risks associated with large model API calls. To test as many large language models (LLMs) as possible, this study utilized both locally deployed models and API calls to access publicly available online models. However, this approach introduces risks such as invocation failure due to compromised API interfaces or credentials, or server crashes. To mitigate the internal threats arising from API call risks, we implemented additional program checkpoints during API invocation. When an API call fails—whether due to network connectivity issues, sensitivity of test data triggering content filters, or other causes—this mechanism allows us to resume the testing procedure from the checkpoint after troubleshooting the fault, thereby avoiding the need for complete retesting.

\paragraph{External validity} External threats primarily center on the specificity of task contexts. For the CoE framework, a significant risk lies in its limited extensibility across diverse task scenarios. Specifically, a CoE framework functioning effectively in one context may fail in others due to distributional shifts in training data. To address these external threats arising from task context specificity, our CoE-Ops framework leverages off-the-shelf general-purpose large models (without specialized training or fine-tuning) combined with advanced prompting techniques. This approach transcends the constraints of specific task contexts, enabling effective routing of expert models across both concrete and abstract domains.

\paragraph{Construct validity} Construct threats primarily center on hallucination issues introduced by the classification model. As our CoE-Ops framework employs a general-purpose large model—without specialized training or fine-tuning—as its classifier, it may exhibit hallucinations when processing high-level tasks. This presents a potential threat to the construct validity of our framework. To mitigate these construct threats, we employ Retrieval-Augmented Generation combined with prompt engineering to reduce hallucination in the classification model.

% \section{Discussion}
% \paragraph{Deployability of CoDER} The CoDER framework achieves enhanced deployability by leveraging general-purpose large language models (LLMs) as task classifiers without requiring fine-tuning or retraining. Furthermore, its expert models can be dynamically adjusted based on benchmark rankings. Thus, the entire system operates exclusively via API calls to both the classifier and expert models, eliminating the need for local model deployment or GPU memory allocation. This design significantly reduces deployment costs and enables practical implementation on edge or mobile devices.

% \paragraph{Future work} The proposed CoDER framework has demonstrated promising results in both task and model scalability. Specifically, without requiring fine-tuning, additional training, or architectural modifications, CoDER achieves effective expert collaboration across diverse DevOps task domains and various combinations of DevOps specialists, thereby enhancing overall model capabilities. However, the current implementation still relies on existing benchmark datasets. For future research directions, we plan to explore automated benchmark construction methodologies to ultimately achieve full automation of the entire workflow. This advancement aims to establish a completely self-contained system that eliminates manual intervention in the benchmarking process.

\section{Conclusion}

% In response to the rapid advancements in software engineering (SE) and large language models (LLMs), coupled with the diverse capabilities demonstrated by expert models across DevOps stages, we propose CoOpERa — a training-free, two-stage expert routing CoE framework tailored for DevOps specialists. This framework replaces domain-specific classifiers in conventional two-stage CoE systems with general-purpose LLMs and integrates prompt engineering to eliminate classifier retraining. Furthermore, we enhance CoOpERa’s capabilities by incorporating Retrieval-Augmented Generation (RAG) techniques. Experimental results demonstrate that CoOpERa effectively combines the strengths of diverse experts to achieve performance improvements across DevOps domains and subtasks. Its training-free design ensures low deployment costs and broad device compatibility. Future work will focus on fully automated benchmark construction to enable end-to-end autonomous collaboration among DevOps experts.

To address the limitations of single AIOps expert models in mastering all DevOps domains and the challenges of ensemble learning in task switching within complex AIOps environments, this paper proposes CoE-Ops, a two-phase expert routing CoE framework based on a general large language model classifier and Retrieval-Augmented Generation. By utilizing the general LLM classifier and prompts, CoE-Ops avoids the need for repeated training or fine-tuning during task scenario transitions, thereby enhancing its task scalability. Furthermore, the incorporation of RAG significantly strengthens its capability in handling tasks with highly abstract scenarios. In future work, we will explore the automated construction of AIOps expert capability rankings to achieve fully automated collaboration among AIOps experts. Additionally, we will integrate this framework with multi-agent systems to establish multi-tiered AIOps expert collaboration.

\end{document}